\documentclass[10pt,twocolumn,letterpaper]{article}

\usepackage[pagenumbers]{iccv} 

\usepackage{adjustbox}
\usepackage{tabularx}
\usepackage{multirow}
\usepackage{colortbl}
\usepackage{xcolor}

\usepackage{amssymb}
\usepackage{pifont}
\newcommand{\cmark}{\ding{51}}%
\newcommand{\xmark}{\ding{55}}%

\newcommand{\ours}{HiERO\xspace}

\newcommand{\pub}[1]{\color{gray}{\scriptsize{({#1})}}}

\DeclareMathOperator*{\mean}{\mathtt{mean}}
\DeclareMathOperator*{\argmax}{arg\,max}

\definecolor{iccvblue}{rgb}{0.21,0.49,0.74}
\usepackage[pagebackref,breaklinks,colorlinks,allcolors=iccvblue]{hyperref}

\title{HiERO: understanding the \underline{hi}erarchy\\ of human b\underline{e}havior enhances \underline{r}easoning on eg\underline{o}centric videos}

\author{Simone Alberto Peirone\\
Politecnico di Torino\\
{\tt\small simone.peirone@polito.it}
\and
Francesca Pistilli\\
Politecnico di Torino\\
{\tt\small francesca.pistilli@polito.it}
\and
Giuseppe Averta\\
Politecnico di Torino\\
{\tt\small giuseppe.averta@polito.it}
}

\begin{document}
\maketitle
\begin{abstract}

Human activities are particularly complex and variable, and this makes challenging for deep learning models to reason about them. 
However, we note that such variability does have an underlying structure, composed of a hierarchy of patterns of related actions. We argue that such structure can emerge naturally from unscripted videos of human activities, and can be leveraged to better reason about their content.
We present \ours, a weakly-supervised method to enrich video segments features with the corresponding hierarchical activity threads.
By aligning video clips with their narrated descriptions, \ours infers contextual, semantic and temporal reasoning with an hierarchical architecture.
We prove the potential of our enriched features with multiple video-text alignment benchmarks (EgoMCQ, EgoNLQ) with minimal additional training, and in zero-shot for procedure learning tasks (EgoProceL and Ego4D Goal-Step). 
Notably, \ours achieves state-of-the-art performance in all the benchmarks, and for procedure learning tasks it outperforms fully-supervised methods by a large margin (\texttt{+}12.5\% F1 on EgoProceL) in zero shot. Our results prove the relevance of using knowledge of the hierarchy of human activities for multiple reasoning tasks in egocentric vision. 
Project page: \href{https://github.com/sapeirone/HiERO}{github.com/sapeirone/HiERO}.
\end{abstract}
\vspace{-10mm}    
\section{Introduction}
\label{sec:intro}

Think about a typical home routine. You enter in the kitchen and grab onions and carrots, chop them, and put them in a pan on the stove with oil. At the same time, you fill a pot with water and put it on the stove. While you wait the water to boil to cook the pasta, you pour some tomatoes in the pan. Zooming out a bit, you can group all these actions into higher-level interleaved activity threads, such as \emph{preparing vegetables} and \emph{cooking pasta}.
Looking at the bigger picture, both of these threads are part of a broader routine like \emph{preparing a meal}, which may overlap with others, such as \emph{washing the dishes}.
Foundational models in egocentric video understanding have long focused mostly on action-level understanding~\cite{omnivore,zhao2023learning,hiervl,pramanick2023egovlpv2,tong2022videomae}, overlooking the inherent hierarchical nature behind human actions~\cite{cooper2006hierarchical,botvinick2009hierarchically,song2024ego4d}. The closest class of approaches that attempts to learn about this compositional structure is Procedure Learning (PL), which assumes that multiple actions concur to form key-steps of long-horizon procedures. However, supervised approaches consider only one level of aggregation, i.e. actions that form key-steps, and require multiple scripted examples of the same procedures to learn from. Conversely, we claim that there is significant value in learning from the hierarchy of human behavior at multiple levels of abstraction. 
Indeed, the richness of human activities lies not only in single actions execution, but more prominently in how these are interconnected at different levels of abstractions. Our intuition is that enriching action features with knowledge of the multiple progressive semantic aggregations they belong to can significantly improve their expressiveness for various reasoning tasks. Interestingly, we believe that such a hierarchical structure can naturally emerge without specific supervision. 
\begin{figure}[t]
    \centering
    \includegraphics[width=1\columnwidth,trim={0 0 0.5cm 0},clip]{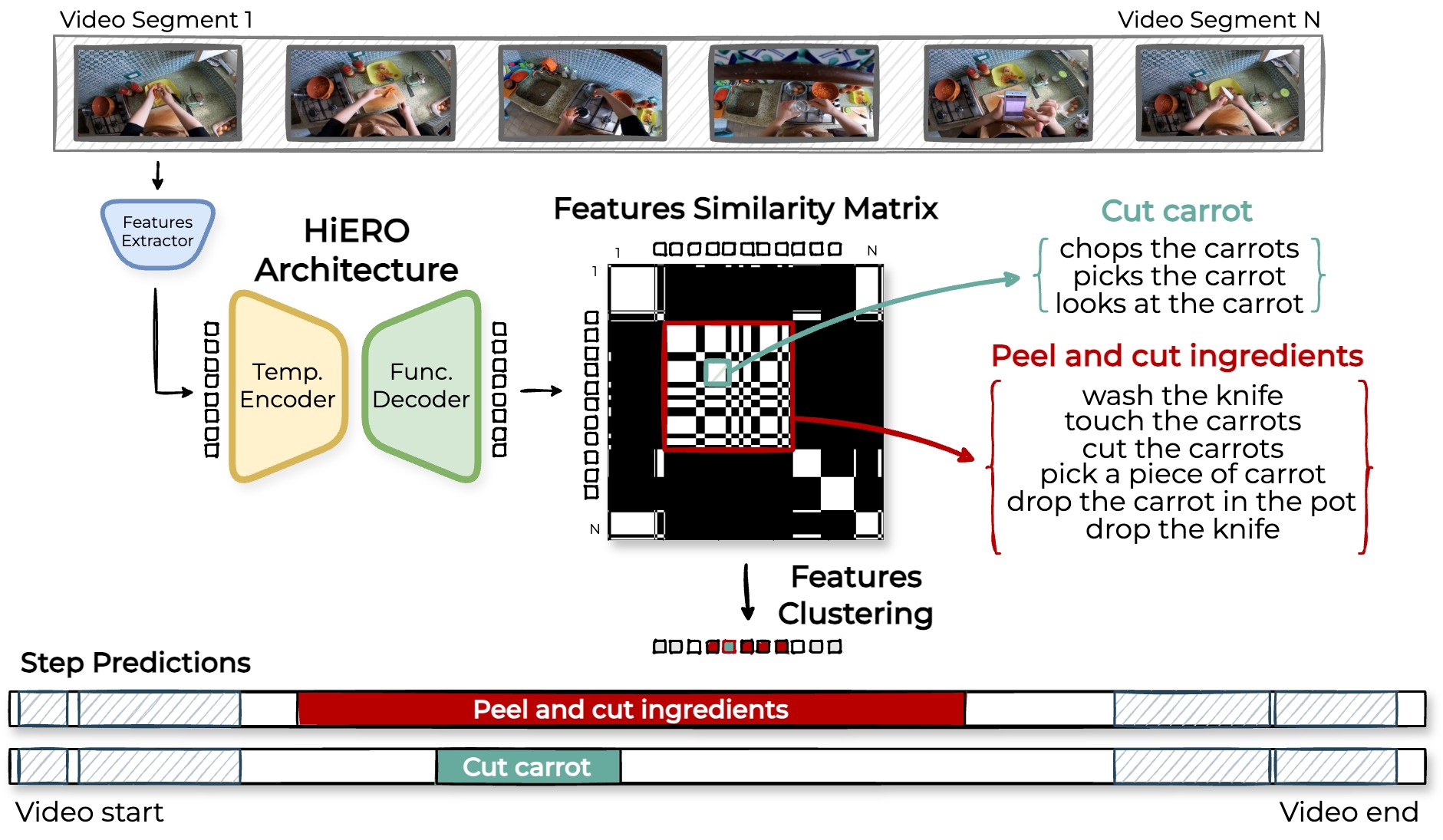}
    \vspace{-5mm}
    \caption{\textbf{Zero-Shot procedure step localization with \ours}. Given a long egocentric video, \ours computes segment-level features that encode the functional dependencies between the actions in the video at different scales. 
    This enables the detection of procedure steps through a simple clustering in feature space.}
    \vspace{-0.5\baselineskip}
    \label{fig:teaser}
\end{figure}
Previous works have shown that even a simple clustering of video segments projected in feature space may be sufficient to identify the high-level activities represented in the video~\cite{bansal2022my,chowdhuryopel,price2022unweavenet,sarfraz2021temporally}. 
However, the choice of the features extractor plays a crucial role in terms of the abstractions that the clustering is able to capture. 
Static biases of video models~\cite{kowal2024quantifying} can generate clusters based on \emph{visual similarity} of the video segments, for example, when two actions occur in the same environment~\cite{nagarajan2020ego}.
Likewise, \emph{semantic similarities} emerge when grouping action segments with similar semantics, \eg, \emph{dicing a carrot} and \emph{slicing an onion}. 
Video-language alignment between short video clips and their corresponding textual descriptions, as seen in EgoVLP~\cite{lin2022egocentric}, results in clip-level representations that capture these similarities by leveraging the proximity of actions in the text space.
At a higher abstraction level, \textbf{functional similarity} groups segments based on their functional objectives, such as identifying all the steps needed to \emph{prepare a meal}. 

With this work, we demonstrate that such functional patterns can emerge naturally from data at different abstraction scales, and can be exploited to enrich features used to solve multiple video understanding tasks in zero shot. 
We model the task as a graph learning problem and represent videos as graphs where nodes correspond to fixed-length video segments and edges reflect the temporal distance between nodes. To preserve and exploit the intrinsic hierarchy of human behavior, we propose to use a hierarchical graph-based representation that provides a strong inductive bias.
This is implemented through a hierarchical architecture inspired by Graph U-Net~\cite{gao2019graph} which we call \ours. 
The model consists of a Temporal Encoder, which gradually aggregates information from nearby nodes within local temporal neighborhoods, and a Function-Aware Decoder which is responsible for discovering strongly connected regions via spectral graph clustering and performs temporal reasoning within each partition separately. 
In this context, activity patterns emerge as strongly connected regions capturing actions that are functionally and temporally related, allowing the model to reason on higher-level activities, see Fig~\ref{fig:teaser}. \ours is trained in a weakly-supervised manner, with the objectives of aligning node features at higher temporal granularity by leveraging video-narration alignment within a temporal window, and guiding clustering at deeper layers to enforce intra-cluster feature proximity.
\ours can perform a wide set of reasoning tasks, including natural language queries, procedure learning, step grounding, and others.
We evaluate the zero-shot transfer of \ours over EgoProceL~\cite{bansal2022my} and Ego4D Goal-Step~\cite{song2024ego4d}, demonstrating remarkable performance compared to fully-supervised models, despite no explicit task-specific training.
By leveraging the inner hierarchical structure of videos, \ours is competitive also with state-of-the-art models on video-text alignment benchmarks with minimal additional training. 
\section{Related works}\label{sec:related}

\noindent\textbf{Long-form understanding.}
Long-form video understanding in egocentric vision requires diverse reasoning abilities to grasp the broader context of human activities~\cite{mangalam2023egoschema,jia2022egotaskqa,islam2024video,peirone2024backpack}, interpret interactions between objects, people, and locations~\cite{price2022unweavenet,nagarajan2020ego,nagarajan2019grounded,goletto2025amego}, and model the procedural nature of human activities~\cite{shen2024progress,seminara2024differentiable,ashutosh2023video}.
Several approaches learn transferable representations for downstream video understanding tasks by aligning short video clips and their corresponding textual narrations~\cite{lin2022egocentric,pramanick2023egovlpv2,hiervl,zhao2023learning}. HierVL~\cite{hiervl} extends this approach by incorporating video-level alignment through summaries. 
Most closely related to ours, \texttt{Paprika}~\cite{zhou2023procedure} exploits supervision from Procedural Knowledge Graphs sourced from wikiHow to develop a set of procedure-aware pre-training objectives.
ProcedureVRL~\cite{zhong2023learning} learns procedure step representations via video-language alignment and a probabilistic model to encode temporal dependencies between individual steps in instructional videos. 
Unlike these approaches, \ours captures long-range functional dependencies between human actions without requiring explicit supervision or instructional video datasets.

\smallskip
\noindent\textbf{Procedure learning.}
PL involves identifying key-steps, \ie, the actions required to complete a task, and predicting their logical order in videos after observing multiple visual demonstrations. 
Supervised approaches \cite{naing2020procedure, zhou2018towards} rely on per-frame key-step annotations across videos, while weakly supervised methods \cite{zhukov2019cross, li2020set, richard2018neuralnetwork} leverage predefined key-step lists
~\cite{lin2022learning,afouras2023ht,mavroudi2023learning,zhou2023procedure}. 
These approaches require extensive annotation efforts, full video observations, or heuristic definitions, making them challenging to scale \cite{elhamifar2020self}.
To mitigate these limitations, self-supervised methods \cite{dvornik2023stepformer,bansal2022my, bansal2024united, chowdhuryopel} have gained attention, as they avoid the need of per-frame annotations. These methods exploit the structured nature of multiple demonstrations of the same task to discover and localize key steps. However, they still rely on the assumption that corresponding actions exist across videos, requiring datasets that contain multiple instances of the same procedure with a shared set of key-steps for alignment. This assumption significantly limits their applicability to real-world, unscripted human activity datasets, restricting their use to well-defined procedural tasks.
In contrast, \ours effectively uncovers meaningful functional threads from unscripted videos without relying on explicit supervision. 

\smallskip
\noindent\textbf{Clustering for vision applications.}
Clustering approaches have been explored to localize objects in the image by looking at densely connected regions of the image~\cite{melas2022deep,wang2023tokencut,wang2024videocutler,shi2000normalized}. 
Self-supervised methods in Procedure Learning~\cite{bansal2022my, bansal2024united, chowdhuryopel} use clustering algorithms to identify procedure steps from features extracted by a self-supervised network trained to align steps across multiple videos of the same task. 
TW-FINCH~\cite{sarfraz2021temporally} tackles unsupervised action segmentation through hierarchical clustering of video segments in feature space, showing that clustering algorithms are surprisingly strong baselines for action segmentation. 
Similarly, Kumar \textit{et al.}~\cite{kumar2022unsupervised} leverages frames clustering as a pretext task, enforcing order-preserving constraints on cluster assignments across videos.
These works are designed for datasets with repetitive and isolated tasks, \eg, 50-Salads~\cite{stein2013combining} and Breakfast~\cite{Kuehne12}. Differently, \ours captures more general functional dependencies between human actions from \emph{in-the-wild} videos, without the need for procedural videos during training.
\section{Method}\label{sec:method}
\begin{figure}[t]
    \centering
    \includegraphics[width=0.95\columnwidth]{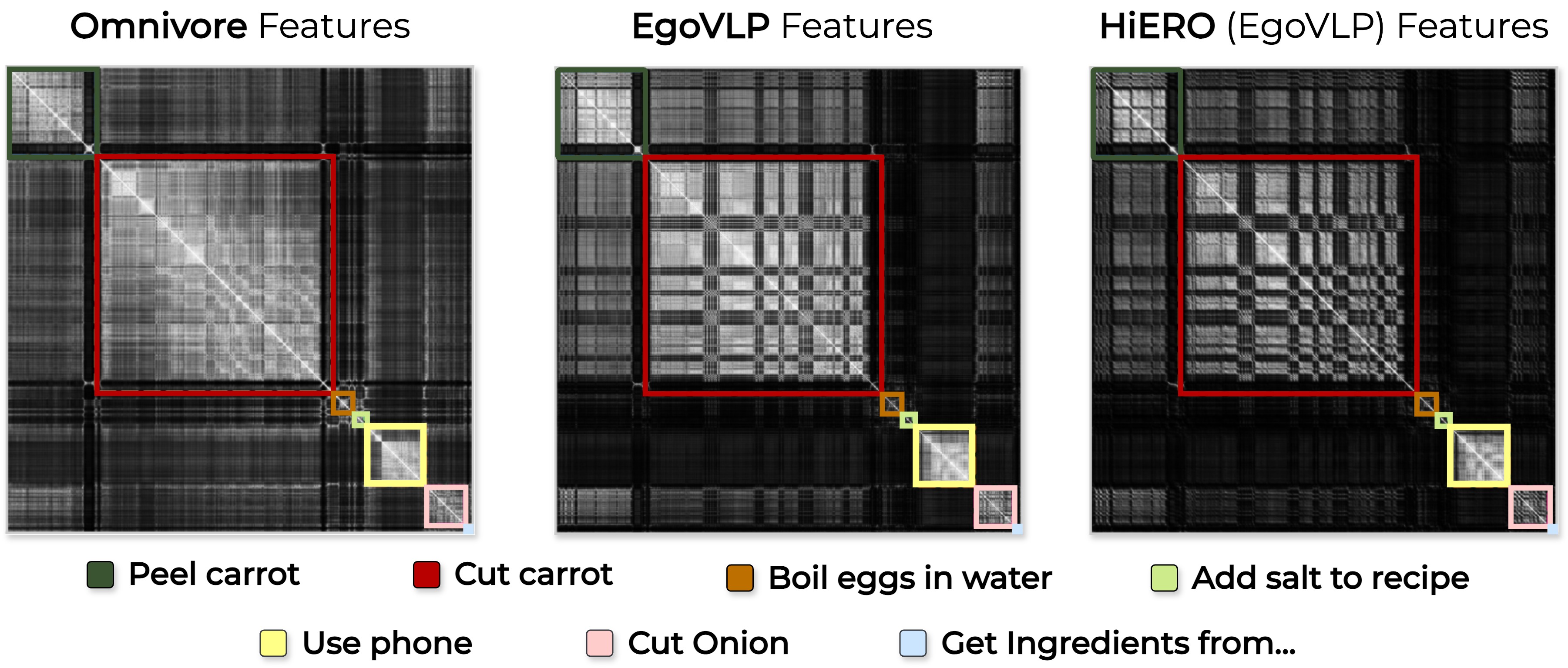} 
    \caption{\textbf{Emergence of step clusters in the features similarity matrix of a video from Ego4D~\cite{ego4d}.} Colored rectangles indicate the ground truth steps. Ideally, we expect high similarity (brighter regions) if two segments represent the same or semantically similar steps, \eg \emph{cut onion} and \emph{cut carrot}. On Omnivore features, this behavior is only partially visible. On EgoVLP features, we observe sharper clusters of temporal segments that are not necessarily close temporally, but represent similar high-level actions. Our approach makes this behavior even more visible.}
    \label{fig:features_similarity}
    \vspace{-1\baselineskip}
\end{figure}

We design \ours based on the intuition that, given a sufficiently large collection of videos capturing human activities in-the-wild, \emph{functional dependencies} between actions naturally emerge as frequently co-occurring patterns directly from observations~\cite{seminara2024differentiable}.
With \ours, we learn a feature space that captures these functional dependencies between actions, \ie, those that frequently co-occur together are close to each other and distant from the others. 
As a result, such space allows related actions to be easily grouped into high-level patterns with a simple clustering operation. 
Our approach represents the video as a graph, in which nodes correspond to short temporal segments, ideally representing one or a few actions, and detects functional threads as regions of this graph whose nodes encode similar actions based on their feature similarity.
Our method builds on spectral graph theory to identify these strongly connected regions of the graph (Sec~\ref{sec:method_background}), learns to detect functional threads by leveraging the natural co-occurrence of human actions in unscripted videos (Sec.~\ref{sec:method_arch}) with the goal of performing different video understanding tasks without additional training (Sec.~\ref{sec:method_zeroshot}).

\smallskip
\noindent\textbf{Functional threads discovery by graph clustering.}\label{sec:step_discovery}
In our context, we define a strongly connected region of the graph as a subset of its nodes showing high \emph{functional similarity}. 
The concept of \emph{similarity} is strongly dependent on the backbone used to compute the node embeddings.
If the backbone was able to map close in feature space the video segments encoding similar actions, \eg, \textit{cutting an onion} and \textit{peeling a carrot}, then these region with high similarity would correspond to high-level \emph{functional threads}, \eg, \textit{preparing the vegetable}. 
To support this intuition, we show in Fig.~\ref{fig:features_similarity} the impact of different backbones on the features similarity matrix.
Omnivore was trained for supervised image and action classification on image and video data respectively. As a result, it focuses mostly on visual similarity between the segments.
On EgoVLP features, some strongly connected regions emerge more clearly, even though the model was trained only with fine-grained narrations supervision.
In the context of procedural videos, these regions may correspond to different steps and substeps of the procedure.
Our approach builds on this intuition to make the unsupervised clustering into high-level functional threads more evident.
The final goal is to partition an input graph $\mathcal{G}$ into a set of $n$ sub-graphs $\{ \mathcal{G}_1, \dots, \mathcal{G}_n\}$, each encoding a different step from the input video.

\subsection{Background: Graph Theory}\label{sec:method_background}
Let $\mathcal{G}$ be an undirected graph with node embeddings $\mathbf{X} \in \mathbb{R}^{N\times D}$, where $N$ is the number of nodes and $D$ the embedding size.
The weighted adjacency matrix $\mathbf{W} \in \mathbb{R}^{N \times N}$ is a nonnegative matrix whose entry $w_{ij}$ is the weight of the edge between nodes $i$ and $j$, while the degree matrix $\mathbf{D} \in \mathbb{R}^{N \times N}$ is a diagonal matrix where $D_{ii} = \sum_{j} \mathbf{W}_{ij}$ is the degree of node $i$. 
The Laplacian matrix $\mathbf{L} = \mathbf{D} - \mathbf{W}$ of the graph is a real symmetric matrix that describes how information flows on the graph: $(\mathbf{L} \mathbf{X})_i = \sum_{j=1}^n w_{ij} (\mathbf{x}_i -  \mathbf{x}_j)$.
The spectral decomposition of the Laplacian Matrix can reveal important topological properties of the graph. 
Most notably, its smallest eigenvalue $\lambda_1$ is zero and the corresponding eigenspace is formed by a set of indicator vectors that identify the connected components of the graph~\cite{von2007tutorial}.

\smallskip
\noindent\textbf{Graph clustering.}
Spectral clustering~\cite{von2007tutorial} groups nodes of the graph such that nodes in each partition are similar to each other.
Unlike other clustering approaches, \eg, K-Means, which require specific assumptions about the data distribution, spectral clustering looks at the connectivity of the graph to groups nodes. 
Given a target number of clusters $K$ to separate, nodes are first projected on the subspace spanned by the eigenvectors of the normalized Laplacian matrix corresponding to its $K$ smallest eigenvalues~\cite{von2007tutorial}.
Then, K-Means is used to cluster the nodes in this subspace.
Given the node embeddings $\mathbf{X}$ of the graph, we build the corresponding similarity matrix $\mathbf{S} \in \mathbb{R}^{N \times N}$~as $S_{ij} = \texttt{exp}\left( \mathbf{x}_i^T \mathbf{x}_j / (\kappa ||\mathbf{x}_i||_2 ||\mathbf{x}_j||_2) \right)$, where $\kappa$ is a temperature parameter.
We define a fully connected \emph{similarity graph} $\mathcal{G}_S$ using $\mathbf{S}$ as adjacency matrix, and define the corresponding normalized Laplacian matrix as $\mathbf{\tilde{L}}_S = \mathbf{I} - \mathbf{D}^{-\frac{1}{2}}\mathbf{S}\mathbf{D}^{-\frac{1}{2}}$,
where $\mathbf{I}$ is the identity matrix and $\mathbf{D}$ is the degree matrix of~$\mathcal{G}_S$.
Then, we find the eigendecomposition of $\mathbf{\tilde{L}}_S$ as $\mathbf{\tilde{L}}_S = \mathbf{U}^T \mathbf{\Lambda} \mathbf{U}$, where $\mathbf{\Lambda} \in \mathbb{R}^{N\times N}$ is a diagonal matrix with the eigenvalues of $\mathbf{\tilde{L}}_S$ on its nonzero entries, and $\mathbf{U} \in \mathbb{R}^{N\times N}$ contains the corresponding eigenvectors on its columns. 
We perform K-Means clustering on the columns of $\mathbf{\tilde{U}} \in \mathbb{R}^{N \times K}$, \ie, the matrix containing the first $K$ eigenvectors on its columns. 
This procedure assigns each node $i$ from $\mathcal{G}$ to one of the $K$ clusters $c_i \in \left[ 1,\dots,K\right]$.

\subsection{The \ours architecture}\label{sec:method_arch}
\begin{figure*}[ht]
    \centering
    \includegraphics[width=0.95\textwidth]{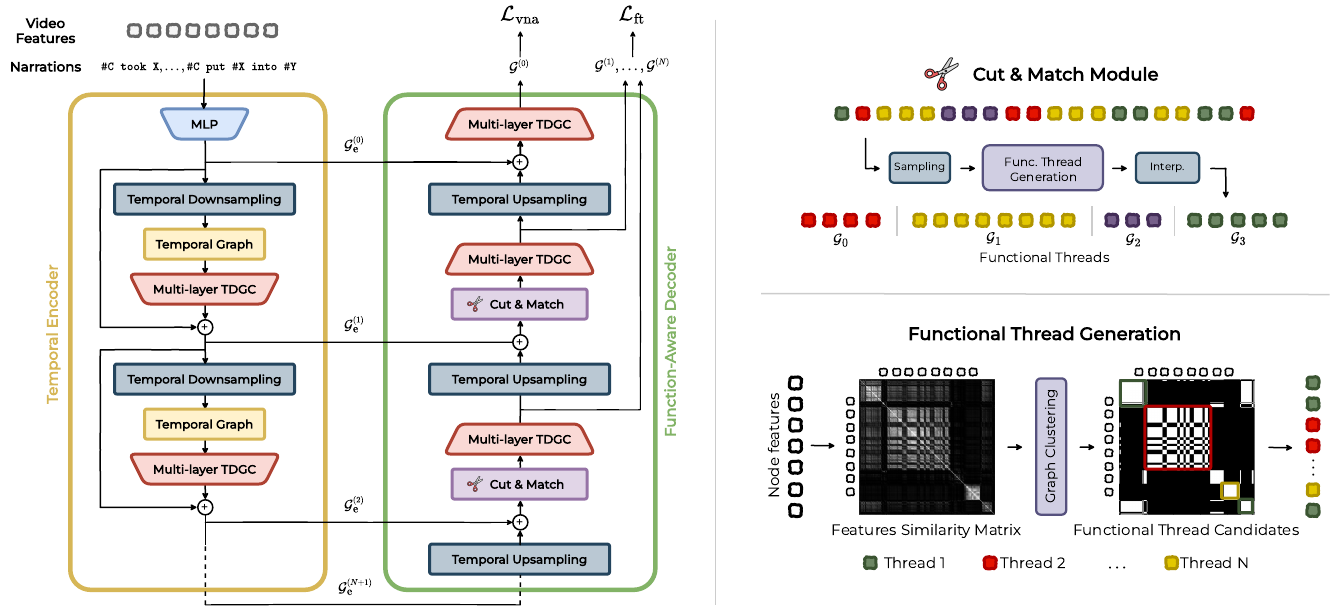}
    \caption{\textbf{Architecture of \ours.} \ours is designed as an encoder-decoder architecture to implement \emph{Function-Aware video-text alignment}. The \textbf{Temporal Encoder} $\mathcal{E}$ performs temporal reasoning on graph representations of the input video at different scales, while the \textbf{Function-Aware Decoder} $\mathcal{D}$ recombines nodes in the video graph by matching segments that represent functional dependencies between the actions (\emph{Cut~\&~Match} module). \ours is trained to align video segments with their corresponding textual narrations at the shallower layer, and to strengthen thread-aware clustering in deeper layers.
    }\label{fig:architecture}
    \vspace{-1\baselineskip}
\end{figure*}
Inspired by previous works in video understanding~\cite{huang2020improving,peirone2025hier}, we encode an input video $\mathcal{V}$ as a \emph{video graph} with $N$ nodes $\mathcal{G} = (\mathbf{X}, \mathcal{E}, \mathbf{p})$, where $\mathbf{X} \in \mathbb{R}^{N \times D}$ is the node embeddings matrix, edge~$e_{ij} \in \mathcal{E}$ connects nodes~$i$ and~$j$ if their temporal distance is smaller than a threshold $\tau$ and the attribute $\mathbf{p} \in \mathbb{R}^{N}$ encodes the temporal position of each node, \ie, its timestamp in seconds.
Each node represents a fixed-length segment of the video and the node embeddings are computed using a video features extractor, such as EgoVLP~\cite{lin2022egocentric}, from the segment frames.
At training time, each video is also associated with a set of narrations, \ie, concise textual descriptions of the actions represented in the video, denoted as $\mathcal{T}_{\mathcal{V}} = \{(n_i, t_i)\}_i$, where $n_i$ and $t_i$ are the textual narration and its corresponding timestamp.
\ours is built as an encoder-decoder architecture inspired by Graph U-Net~\cite{gao2019graph}.
The two branches share the same components but serve different roles. 
The \emph{Temporal Encoder}~$\mathcal{E}$ implements local temporal reasoning, hierarchically aggregating information between temporally close segments, while the \emph{Function-Aware Decoder}~$\mathcal{D}$ extends temporal reasoning to nodes that may be temporally distant but functionally similar, by connecting nodes belonging to the same thread.
The architecture of \ours is presented in~Fig.~\ref{fig:architecture}.

\smallskip
\noindent\textbf{Temporal Encoder.}
The \emph{Temporal Encoder} $\mathcal{E}$ is implemented as a stack of $N_l$ GNN-based blocks with temporal subsampling operations to map the input video graph $\mathcal{G}^{(0)}$ to a set of temporally coarsened representations:
\begin{equation}
    \mathcal{E}: \mathcal{G}^{(0)} \to \{\mathcal{G}^{(1)}_e, \mathcal{G}^{(2)}_e, \dots, \mathcal{G}^{(N_l)}_e\}.
\end{equation}
At the stage of the encoder at depth $l$, the temporal neighborhood of each node is defined as the set of all the nodes within a certain temporal distance $d$, adjusted for the depth $i$ of the encoder stage. 
Each stage is composed of multiple TDGC~\cite{peirone2025hier} layers that implement temporal reasoning on the graph by combining the embedding of node $i$ with a learnable projection of its neighbors $\mathcal{N}(i)$:
\begin{align}
    \mathbf{x}_j^{'}     & = \mathtt{MLP}\left(\mathbf{x}_j^{l}\right) = \phi(\mathbf{W}_n^T \mathbf{x}_j^{l} + \mathbf{b}_n),                                                  \\
    \mathbf{x}_i^{l+1} & = \mathbf{W}^T_r\mathbf{x}_i^{l} + \mean_{j \in \mathcal{N}(i)} \left( s_{ij} ( \mathbf{w}_{ij}  \odot \mathbf{x}_j^{'}) \right) + \mathbf{b}_r,
\end{align}
where $\mathbf{W}_n^T$ and $\mathbf{W}^T_r$ are learnable projection matrices, $\mathbf{b}_n$ and $\mathbf{b}_r$ are bias terms. $s_{ij}$ and $\mathbf{w}_{ij}$ are used to rescale the contribution of each node depending on its temporal distance and are computed as:
\begin{equation}
    s_{ij} = \mathtt{sign}(\mathbf{p}_{[i]}^{l} - \mathbf{p}_{[j]}^{l}), \;\;\; \mathbf{w}_{ij} = \mathtt{MLP}(|\mathbf{p}_{[i]}^{l} - \mathbf{p}_{[j]}^{l}|).
\end{equation}
Then, the nodes are subsampled to halve the temporal resolution of the graph and obtain $\mathcal{G}^{(l + 1)}$, which is fed to the next layer of the encoder.
Therefore, the encoder progressively extends the temporal context of the nodes, regardless of whether the actions performed are related or not.

\smallskip
\noindent\textbf{Function-Aware Decoder.}
The \emph{Function-Aware Decoder}~$\mathcal{D}$ shares the same architecture of the encoder with one significant difference: instead of implementing message passing on the local temporal neighborhood of the nodes, each decoder stage first groups the graph nodes based on their functional similarity, \ie, whether they represent functionally similar actions, and then implements temporal reasoning on each group separately. 
This procedure connects nodes that may be temporally distant but encode similar actions (\emph{functional threads}), allowing the model to reason about long-term patterns not necessarily connected in time.
The training process of \ours (Sec.~\ref{sec:method_training}) explicitly pushes nodes that are assigned to the same cluster to be close in the features space and far from nodes assigned to other clusters. 

First, each stage $l$ of the decoder takes graph $\mathcal{G}^{l}_e$ from the corresponding temporal encoder stage via a lateral connection and the output of the upper layer of the decoder $\mathcal{G}^{l+1}_d$. 
The node features of $\mathcal{G}^{l+1}_d$ are then interpolated to match the temporal resolution of $\mathcal{G}^{l}_e$ and the two contributions are summed together.
The resulting graph $\tilde{\mathcal{G}}^{l+1}_d$ is then fed to the \emph{Cut~\&~Match} module (Fig.~\ref{fig:architecture}), which partitions the graph into a set of $K$ smaller graphs $\{\tilde{\mathcal{G}}^{l+1}_{d,1}, \dots, \tilde{\mathcal{G}}^{l+1}_{d,K}\}$, each corresponding to a group of functionally similar nodes, following the process described in Sec.~\ref{sec:method_background}. 
The tensor $\mathbf{c}^{l+1} \in [1, \dots, K ]^n$ encodes the cluster assignment for each of the $K$ sub-graphs obtained from $\tilde{\mathcal{G}}^{l+1}_d$.
After this process, nodes that correspond to far apart segments of the video may be clustered together.
We use TDGC to perform temporal reasoning into each partition separately and map the nodes back to the original graph to obtain $\mathcal{G}^{l+1}_d$, which becomes the input of the next decoder layer.
As the number of nodes in the graph may grow rapidly with the size of the graph, we subsample the graph to a fixed and smaller size before processing with the \emph{Cut~\&~Match} module, compute the cluster assignments and propagate them back to the original nodes using a \texttt{1-NN} approach. More details on this approach in the appendix.

\subsubsection{Training \ours}\label{sec:method_training}
We train \ours to map video segments representing co-occurring actions close in the feature space (\emph{video-narrations alignment loss} $\mathcal{L}_{vna}$) and to detect functional threads not necessarily close in time (\emph{functional threads loss} $\mathcal{L}_{ft}$), for example in the case of interleaved activities, without using any specific supervision other than textual narrations.
\ours is trained with a combination of the two losses $\mathcal{L} = \mathcal{L}_{vna} + \mathcal{L}_{ft}$.

\smallskip
\noindent\textbf{Video-narrations alignment.} 
The \emph{video-narrations alignment loss} $\mathcal{L}_{vna}$ encourages the network to map closer in the features space actions that typically occur together. 
$\mathcal{L}_{vna}$ is inspired by previous works in video-language pretraining~\cite{lin2022egocentric,pramanick2023egovlpv2,zhao2023learning} and is defined as a contrastive loss that pushes the node embeddings close to the text embeddings of the narrations that fall within a certain temporal window around the timestamp associated to the node (\emph{positives}), while pushing apart narrations outside the window or appearing in other videos in the batch (\emph{negatives}). 
Unlike previous works, like EgoVLP~\cite{lin2022egocentric}, which align each temporal segment with one single narration from the same window, ignoring the temporal context in which actions occur, our approach explicitly considers the co-occurrence of multiple actions in the temporal window covered by the node, making the embedding more context-aware. As a result, usually co-occurring actions have more similar embeddings that could be clustered more easily into high-level patterns.

Given a batch of B graphs $\{\mathcal{G}_1,\dots,\mathcal{G}_B\}$ with their corresponding narrations $\{\mathcal{T}_1,\dots,\mathcal{T}_B\}$, we define $\mathbf{V}_i \in \mathbb{R}^{N \times D_v}$ as the node embeddings of graph $\mathcal{G}_i$ at the output of the last decoder layer. 
The set of positive narrations for node $j$ from graph $\mathcal{G}_i$ is defined~as $\mathcal{P}_{j} = \{ (n, t) \in \mathcal{T}_i \;\text{s.t.}\; |p - t| \le 2^{\alpha} \}$, where $p$ is the timestamp associated to the node. 
Similarly, the negatives are defined as the narrations outside the window associated to the node or from other videos in the batch: $\mathcal{N}_j = \{ (n, t) \in \mathcal{T}_i \;\text{s.t.}\; 2^{\alpha} < |p - t| \le 2^{\beta}\} \;\cap\; \mathcal{T}_{k, k \ne i}$. $\alpha$ and $\beta$ control the size of the alignment window for positives and negatives sampling.
Formally, the loss $\mathcal{L}_{vna}$ is defined as the sum of two symmetric contributions for \textit{video-to-text} ($\mathcal{L}_{v2t}$) and \textit{text-to-video} ($\mathcal{L}_{t2v}$) alignment:
\begin{equation}
    \hspace{-1.5mm}\mathcal{L}_{v2t}\hspace{-.5mm}=\hspace{-.5mm}\frac{1}{B} \sum_{\mathbf{v}_j} \frac{\sum_{n\in\mathcal{P}_j } \text{exp} (h_v(\mathbf{v}_j)^T h_t(\mathcal{F}(n))/\tau)}{\sum_{n\in\mathcal{P}_j \cup \mathcal{N}_j} \text{exp} (h_v(\mathbf{v}_j)^T h_t(\mathcal{F}(n))/\tau)},
\end{equation}
where $h_v$ and $h_t$ are linear projections followed by L2-normalization to map the visual and textual embeddings in the same features space for alignment, $\mathcal{F}$ is a text features extractor, \eg, BERT, and $\tau$ is a temperature parameter.
The \textit{text-to-video} loss ($\mathcal{L}_{t2v}$) is symmetrically defined in the same way.

\smallskip
\noindent\textbf{Functional threads loss.} 
Aligning the visual embeddings from larger temporal windows to their corresponding textual descriptions is more difficult.
Using narrations is impractical as they are too fine-grained and the number of positive and negatives samples would grow rapidly with the depth of the network and the size of the alignment window. 
Other forms of \emph{high-level} supervision, \eg, video summaries, require huge annotation efforts.
Instead, we apply video-narrations alignment only on the output of the decoder and introduce a contrastive regularization objective to make features at deeper layers belonging to the same functional thread more similar to each other. 
The \emph{functional threads loss} $\mathcal{L}_{ft}$ leverages the graph partition assignments from the \emph{Cut~\&~Match} modules in the decoder and pushes closer to each other samples that are assigned to the same cluster, while pushing away samples from other clusters.
Specifically, given the node embeddings $\mathbf{V}_i^{l} \in \mathbb{R}^{n \times D_v}$ at the output of the decoder for graph $\mathcal{G}_i$ with $n$ nodes at depth $l$, the $\mathcal{L}_{ft}$ is defined as:
\begin{equation}
\hspace{-2mm}\mathcal{L}_{ft} = \sum_{k=1}^K \sum_{\substack{i = 1 \\ c_i = k}}^n \sum_{\substack{j = 1 \\ c_i = c_j}}^n \frac{ \text{exp} (h_v(\mathbf{v}_i)^T h_v(\mathbf{v}_j)/\tau)}{\sum_{j^{'}=1}^n \text{exp} (h_v(\mathbf{v}_i)^T h_v(\mathbf{v}_{j^{'}})/\tau)},
\end{equation}
where $c_i$ represents the cluster assignment of node $i$.

\subsection{Zero-shot procedural tasks}\label{sec:method_zeroshot}
After training, \ours can detect candidate procedure steps by clustering the output of the decoder at different granularities.
This enables our approach to address different procedure learning tasks, including the segmentation of all the steps in the video (\textbf{\emph{procedure learning}}), the temporal grounding of a step given its free-form textual description (\textbf{\emph{step grounding}}) and the localization and classification of all the steps and sub-steps in a video (\textbf{\emph{step localization}}), without any additional training.

Given the node embeddings $\mathbf{V}_i^{l} \in \mathbb{R}^{n \times D_v}$ of graph $\mathcal{G}_i$ with $n$ nodes at depth $l$, we apply the clustering method proposed in Sec.~\ref{sec:step_discovery} to assign each node in the graph to one out of $K$ possible clusters: $\mathbf{c}^l \in [1,\dots,K]^n$. 
The cluster assignments are then upsampled to match the frame rate of the input video $\mathbf{c} = \mathtt{UP}(\mathbf{c}^l)$.
For \emph{procedure learning}, $\mathbf{c}$ represents the step assignments for each segment.
For other tasks, we map the output features of the decoder using $h_v$, apply the clustering algorithm and aggregate features from consecutive segments that are assigned to the same step to obtain a set of $M$ candidate step embeddings $\{\mathbf{F}_1, \dots, \mathbf{F}_M\}$ with $\mathbf{F}_i \in \mathbb{R}^{D_v}$. 
Short candidate segments are discarded as background. 
For \emph{step localization}, given a textual taxonomy consisting of $S$ step labels and the corresponding textual embeddings $\mathbf{T} \in \mathbb{R}^{S \times D_t}$, we assign each step candidate the label $y_i$ that maximizes the cosine similarity between the average visual features of the segment $\mathbf{f}_i$ and the step label embedding: $y_i = \argmax_j \mathbf{f}_i^T \mathbf{t}_j/ \|\mathbf{f}_i\| \|\mathbf{t}_j\|$.
For \emph{step grounding}, we extract an embedding $\mathbf{t}$ from the textual query and select the candidate steps based on cosine similarity between their visual features and the query embedding.
More details on downstream tasks implementation are reported in the appendix.

\section{Experiments}\label{sec:experiments}
We train \ours on EgoClip~\cite{lin2022egocentric}, a curated set of 3.8M clip-text pairs obtained from Ego4D textual narrations, using pre-extracted features from several backbones, \ie, Omnivore~\cite{omnivore}, EgoVLP~\cite{lin2022egocentric} and \textsc{LaViLa}~\cite{zhao2023learning}, showing that \ours can be easily applied to different backbones. 
For EgoVLP and \textsc{LaViLa} we reuse their text encoders when training \ours, while for Omnivore we start from a pre-trained DistillBERT~\cite{sanh2019distilbert} model and fine-tune it during training. We train \ours for 15 epochs, using batch size 8 and learning rate $1\times10^{-5}$ with linear warmup for the first 5 epochs and a cosine annealing schedule. Training takes less than 20 GPU hours.
More details in the appendix.

\smallskip
\noindent\textbf{Evaluation benchmarks.} 
We evaluate our approach on several egocentric vision benchmarks to validate its effectiveness in different scenarios.
Specifically, we validate the video-text alignment components of \ours on \textbf{EgoMCQ}~\cite{lin2022egocentric}, a set of 39K \emph{text-to-video} multiple-choice questions derived from Ego4D narrations, and \textbf{EgoNLQ}, a natural language queries benchmark that aims to localize the segment of a video (start and end timestamps) answering a given textual query. 
For Procedure Learning, we evaluate \ours on \textbf{EgoProceL}~\cite{bansal2022my}, a large scale benchmark with 62 hours of procedural videos from a set of 16 different tasks, and on the Step Grounding and Step Localization tasks from \textbf{Goal-Step}~\cite{song2024ego4d}, a subset of Ego4D featuring procedure annotations from a taxonomy of 514 fine-grained steps and substeps.
The design of \ours allows to address most of these tasks in a completely \emph{zero-shot} setting.

\subsection{Quantitative Results}
\begin{table}[!t]
    \centering

    \scriptsize
    \setlength{\tabcolsep}{4pt}
    \begin{tabularx}{\columnwidth}{@{}Xcccccc@{}}
        \toprule

        \multirow{2}{*}{\bf Method}                           & \multicolumn{2}{c}{\bf EgoMCQ}    & \multicolumn{4}{c}{\bf EgoNLQ}                                                                                           \\

        \cmidrule{2-7}

                                                              & \multicolumn{2}{c}{Accuracy (\%)} & \multicolumn{2}{c}{mIOU@0.3}   & \multicolumn{2}{c}{mIOU@0.5}                                                            \\

                                                              & Inter                             & Intra                          & R@1                          & R@5               & R@1              & R@5               \\

        \midrule

        Omnivore~\cite{omnivore}$^\dagger$\pub{CVPR'22}       & $-$                               & $-$                            & 6.56                         & 12.55             & 3.59             & 7.90              \\
        SlowFast~\cite{slowfast}~\pub{ICCV'19}                & $-$                               & $-$                            & 5.45                         & 10.74             & 3.12             & 6.63              \\
        EgoVLP~\cite{lin2022egocentric}~\pub{NIPS'22}         & 90.6                              & 57.2                           & 10.84                        & 18.84             & 6.81             & 13.45             \\
        HierVL~\cite{hiervl}~\pub{CVPR'23}                    & 90.5                              & 52.4                           & $-$                          & $-$               & $-$              & $-$               \\
        \textsc{LaViLa}~\cite{zhao2023learning}~\pub{CVPR'23} & \underline{94.5}                     & \underline{63.1}               & 12.05                        & \underline{22.38} & 7.43             & \underline{15.44}             \\
        EgoVLPv2~\cite{pramanick2023egovlpv2}~\pub{ICCV'23}   & 91.0                              & 60.9                           & \underline{12.95}            & \textbf{23.80}    & \underline{7.91} & \textbf{16.11} \\

        \midrule

        \textbf{Ours (Omnivore)}                                       & 90.1                              & 53.4                           & 10.27                        & 18.20             & 6.01             & 12.52             \\
        \textbf{Ours (EgoVLP)}                                         & \underline{91.6}                  & 59.6                           & 11.41 & 19.67 & 7.05 & 13.91             \\
        \textbf{Ours (\textsc{LaViLa})}               & \textbf{94.6}                     & \textbf{64.4}                  & \textbf{13.35}               & 21.12             & \textbf{8.08}    & 15.31    \\

        \bottomrule
    \end{tabularx}
    \vspace{-3mm}
    \caption{\textbf{Results on EgoMCQ and EgoNLQ's validation set}, using VSLNet~\cite{zhang2020span} as grounding head for the latter. $^\dagger$Reproduced.}
    \vspace{-\baselineskip}
    \label{tab:egomcq_nlq}
    \vspace{-0.2cm}
\end{table}
\begin{table*}[t]
 \centering
 \scriptsize
 \setlength{\tabcolsep}{6.5pt}
 \begin{tabularx}{\textwidth}{@{}Xcc|cccccccccccc@{}}
 \toprule
 \multirow{2}{*}{\bf Method} & \multicolumn{2}{c}{\textbf{Average}} & \multicolumn{2}{c}{\textbf{CMU-MMAC}~\cite{cmu_mmac}} & \multicolumn{2}{c}{\textbf{EGTEA}~\cite{li2018eye}} & \multicolumn{2}{c}{\textbf{MECCANO}~\cite{ragusa_MECCANO_2023}} & \multicolumn{2}{c}{\textbf{EPIC-Tents}~\cite{epic_tent}} & \multicolumn{2}{c}{\textbf{PC Ass.}~\cite{bansal2022my}} & \multicolumn{2}{c}{\textbf{PC Disass.}~\cite{bansal2022my}} \\
 \cmidrule(lr){2-3}\cmidrule(lr){4-5}\cmidrule(lr){6-7}\cmidrule(lr){8-9}\cmidrule(lr){10-11}\cmidrule(lr){12-13}\cmidrule(l){14-15}
 \multicolumn{1}{c}{} & F1 & \multicolumn{1}{c}{IoU} & F1 & IoU & F1 & IoU & F1 & IoU & F1 & IoU & F1 & \multicolumn{1}{c}{IoU} & F1 & IoU \\
 \midrule
 Random~\cite{chowdhuryopel} \pub{NeurIPS'24} & 14.8 & 6.1 & 15.7 & 5.9 & 15.3 & 4.6 & 13.4 & 5.3 & 14.1 & 6.5 & 15.1 & 7.2 & 15.3 & 7.1 \\
 CnC~\cite{bansal2022my} \pub{ECCV'22} & 22.0 & 10.7 & 22.7 & 11.1 & 21.7 & 9.5 & 18.1 & 7.8 & 17.2 & 8.3 & 25.1 & 12.8 & 27.0 & 14.8 \\
 GPL-2D~\cite{bansal2024united} \pub{WACV'24} & 22.0 & 11.9 & 21.8 & 11.7 & 23.6 & 14.3 & 18.0 & 8.4 & 17.4 & 8.5 & 24.0 & 12.6 & 27.4 & 15.9 \\
 GPL~\cite{bansal2024united} \pub{WACV'24} & 25.6 & 13.9 & 31.7 & 17.9 & 27.1 & 16.0 & 20.7 & 10.0 & 19.8 & 9.1 & 27.5 & 15.2 & 26.7 & 15.2 \\
 OPEL~\cite{chowdhuryopel} \pub{NeurIPS'24} & 32.0 & 16.3 & 36.5 & 18.8 & 29.5 & 13.2 & 39.2 & 20.2 & 20.7 & 10.6 & 33.7 & 17.9 & 32.2 & 16.9 \\
 \midrule
 \midrule
 Omnivore & 39.1 & 22.0 & 44.7 & 26.8 & \underline{37.1} & 19.2 & 36.0 & 19.0 & \underline{40.8} & \underline{21.9} & 35.7 & 21.5 & 40.3 & 23.5 \\
 \textbf{Ours (Omnivore)} & \underline{44.0} & \underline{24.5} & 47.2 & 27.7 & \textbf{39.7} & \textbf{19.9} & \textbf{41.6} & \textbf{22.1} & \textbf{45.3} & \textbf{24.3} & \underline{43.7} & \underline{25.1} & \underline{46.3} & \underline{27.9}\\
 \midrule
 EgoVLP & 40.0 & 21.9 & \underline{49.2} & \underline{31.0} & 36.6 & 18.3 & 33.1 & 16.1 & 37.4 & 19.2 & 38.2 & 20.8 & 45.4 & 25.6 \\
 \textbf{Ours (EgoVLP)} & \textbf{44.5} & \textbf{25.3} & \textbf{53.5} & \textbf{34.0} & \textbf{39.7} & \underline{19.6} & \underline{39.8} & \underline{20.3} & 39.0 & 20.3 & \textbf{44.9} & \textbf{25.6} & \textbf{49.9} & \textbf{32.1} \\
 \bottomrule
 \end{tabularx}
 \vspace{-3mm}
 \caption{\textbf{Comparison with the state-of-the-art on the EgoProceL benchmark~\cite{bansal2022my}}.
 Performance is evaluated in terms of F1 score and IoU w.r.t. ground truth key-steps, using a fixed number of predicted key-steps ($k=7$) for a fair comparison to the previous approaches.}
 \vspace{-\baselineskip}
 \label{tab:egoprocel_main}
 \vspace{-0.1cm}
\end{table*}

\subsubsection{Video-Text Alignment on EgoMCQ and EgoNLQ}
We evaluate \ours on EgoMCQ~\cite{lin2022egocentric} and EgoNLQ~\cite{ego4d} to validate its video-text alignment capabilities and to show that reasoning on functional threads at different scales can support various video understanding tasks (Table~\ref{tab:egomcq_nlq}). 
EgoMCQ is a multiple-choice \emph{text-to-video} retrieval task where the goal is to select the video clip that matches a given textual description among five possible candidates. 
Results are measured in terms on \emph{inter} (options are from different videos) and \emph{intra} accuracy (options are from the same video).
EgoNLQ aims at localizing the temporal segment of a video that answers a textual query, \eg, \textit{Where did I put X?} or \textit{Where is object X before / after event Y?}. 
These queries require strong temporal and causal understanding of the interactions between different objects and actions in the video.
Performance is measured with Recall at different IoU thresholds between the predicted and the ground truth segments.
For this task, we follow previous approaches~\cite{ego4d,lin2022egocentric,zhao2023learning,pramanick2023egovlpv2} and train a VSLNet~\cite{zhang2020span} grounding head on top of the features at the output of the decoder of \ours.

Our window-based alignment loss encourages \ours to learn functional dependencies between actions, while clustering groups together similar actions at different scales and over a long temporal horizon.
Together, these objectives are effective to discriminate between similar short-term actions, which is critical for EgoMCQ, as well as to capture long-range causal and temporal dependencies in the video, which is essential for EgoNLQ. 
Unlike other backbones that extract features from a short temporal window and rely entirely on the grounding head for high-level reasoning, our features inherently capture a broader semantic understanding of the video.
In both benchmarks, \ours significantly improves the SOTA, regardless of the features extraction backbone ($\texttt{+}1.3\%$ on intra accuracy on EgoMCQ and Top-1 Recall at IoU = 0.3 on EgoNLQ when using \textsc{LaViLa} features).
Remarkably, \ours achieves good results even with Omnivore features, despite not being trained end-to-end on~Ego4D.

\subsubsection{Procedure Learning on EgoProceL}
We evaluate \ours on EgoProceL~\cite{bansal2022my} in \emph{zero-shot}, using visual features extracted from the Omnivore and EgoVLP backbones.
Following the original evaluation protocol~\cite{bansal2022my}, we compute frame-wise step assignments and  match them with the ground truth using the Hungarian algorithm~\cite{kuhn1955hungarian}. Performance is measured in terms of F1 score and IoU with respect to the ground truth key-steps. More details in the appendix.
Compared to previous works in this setting that are based on matching visual segments between pairs of videos representing the same task, \eg, CnC~\cite{bansal2022my}, GPL~\cite{bansal2022my} and OPEL~\cite{chowdhuryopel}, our approach is fundamentally different and does not require any additional supervision. 
Indeed, we evaluate on this benchmark the ability of \ours to group together parts of the video that correspond to the same high-level activity by leveraging their functional similarity, even though it was not trained explicitly to identify procedure steps inside a video.
We compare \ours with the SOTA in Table~\ref{tab:egoprocel_main}, using a fixed number of key-steps to predict ($k=7$) for a fair comparison with previous approaches that share the same assumption.
Using our clustering approach in combination with Omnivore and EgoVLP features is already particularly effective in detecting the procedure steps ($\texttt{+}7.1\%$ and $\texttt{+}8.0\%$ respectively compared to the previous state-of-the-art OPEL~\cite{chowdhuryopel}), supporting the intuition that steps can emerge as clusters of similar actions~\cite{sarfraz2021temporally}.
\ours significantly improves over these baselines ($\texttt{+}4.9\%$ and $\texttt{+}4.5\%$ respectively), showing that i) procedure steps can emerge by actions clustering without the need of specific supervision, ii) \ours can generalize to novel procedural tasks that were not present in Ego4D.
We present an in depth comparison of our baselines and OPEL in the appendix.

\begin{figure*}[ht]
\centering
 \begin{subfigure}{0.475\textwidth}
     \centering
     \includegraphics[width=\textwidth]{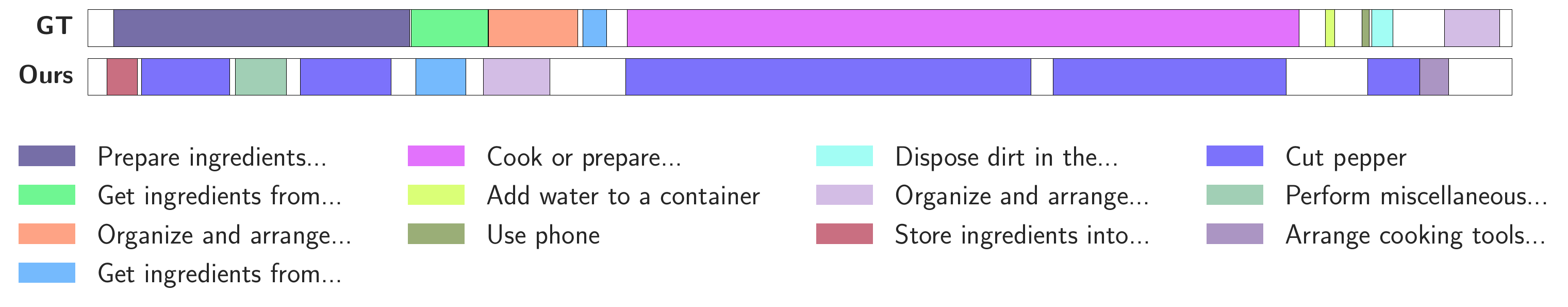}
     \caption{Video 1 (\texttt{0f07958c-04e3-4be9-9118-f3313c4e183e})}
     \label{fig:zs_segm_a}
 \end{subfigure}
 \hspace{5mm}
 \begin{subfigure}{0.475\textwidth}
 \centering
     \includegraphics[width=\textwidth]{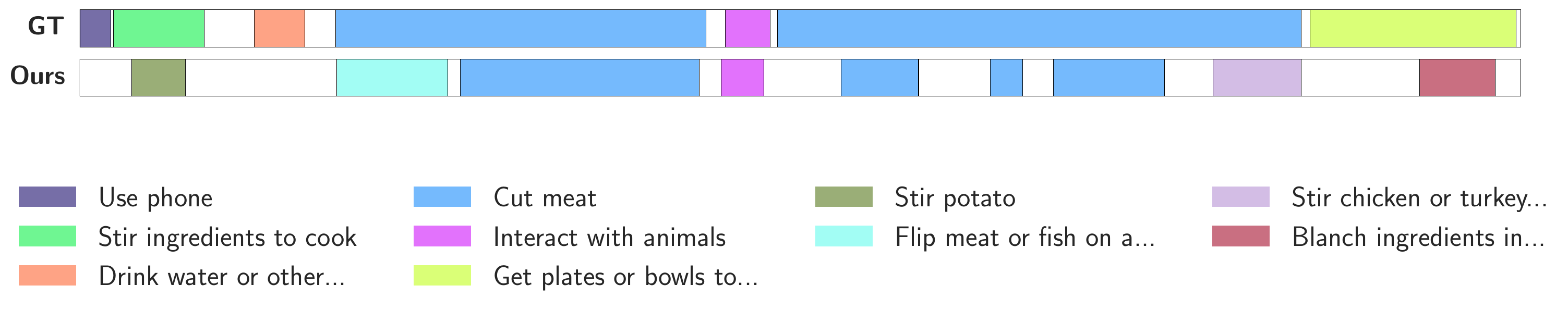}
     \caption{Video 2 (\texttt{4bddae9e-8ffb-4a03-9421-adf6268d91b6})}
     \label{fig:zs_segm_b}
 \end{subfigure}
 
 \smallskip
 \begin{subfigure}{0.475\textwidth}
     \centering
     \includegraphics[width=\textwidth]{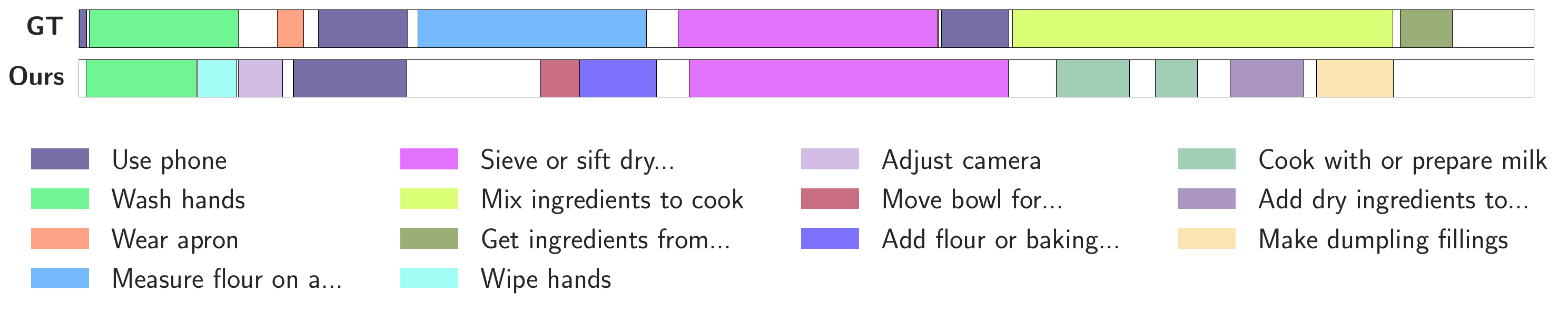}
     \caption{Video 3 (\texttt{603a427f-9191-4ca4-a1b0-dd3c5e7fda70})}
     \label{fig:zs_segm_c}
 \end{subfigure}
 \hspace{5mm}
 \begin{subfigure}{0.475\textwidth}
     \centering
     \includegraphics[width=\textwidth]{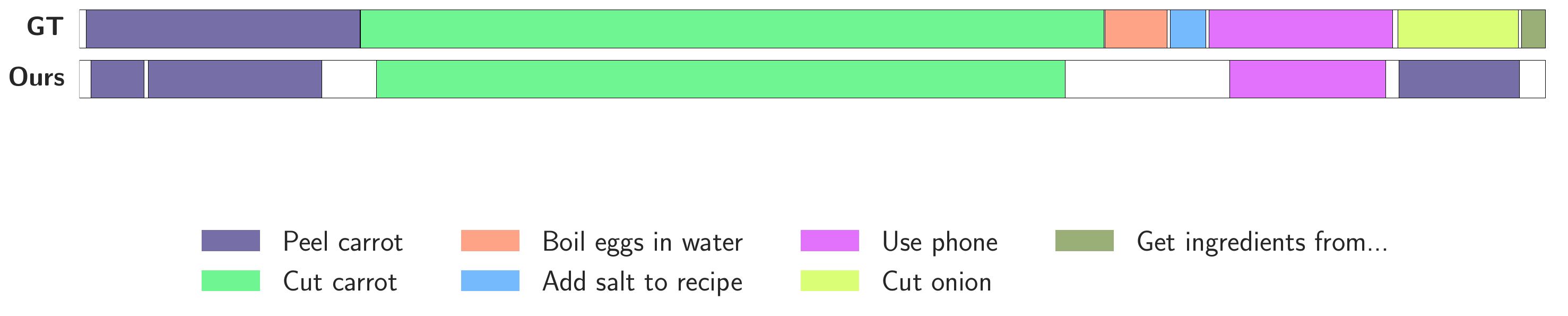}
     \caption{Video 4 (\texttt{64b355f3-ef49-4990-8622-9e9eef68b495})}
     \label{fig:zs_segm_d}
 \end{subfigure}
\vspace{-2mm}
 \caption{\textbf{Zero-Shot Localization results on Ego4D Goal-Step, showing some of the \ours's success and failure cases.} We observe that many failure cases of \ours are related to the ambiguous granularity of the step annotations in the dataset. In Fig.~\ref{fig:zs_segm_a}, \ours confuses the step \emph{Cook or prepare the vegetables} with the closely related \emph{Cut the pepper}. In Fig.~\ref{fig:zs_segm_c}, \ours correctly identifies many steps but confuses \emph{Mix ingredients to cook} with some of its possible sub-steps, \eg, \emph{Cook with or prepare milk}.
 }
 \vspace{-0.5\baselineskip}
 \label{fig:zs_segm}
\end{figure*}

\subsubsection{Step Grounding and Localization on Goal-Step}
\begin{table}
\centering
\scriptsize
\setlength{\tabcolsep}{5pt}
\begin{tabularx}{\columnwidth}{@{}Xccccc@{}}
\toprule
\multirow{2}{*}{\bf Method}& \multirow{2}{*}{\bf Approach} & \multicolumn{2}{c}{\bf mIoU@0.3} & \multicolumn{2}{c}{\bf mIoU@0.5}\\
\cmidrule(lr){3-4}\cmidrule(lr){5-6}
 & & R@1 & R@5 & R@1 & R@5 \\
\midrule
Omnivore~\cite{song2024ego4d} & Supervised & 12.02 & 19.99 & 7.71 & 14.17 \\
\bf Ours (Omnivore) & Supervised & 13.02 & 21.81 & 8.59 & 15.98 \\
\midrule 
EgoVLP & Supervised & \underline{15.43} & \underline{25.91} & \underline{10.95} & \underline{19.77}\\
\bf Ours (EgoVLP) & Supervised & \textbf{15.64} & \textbf{26.01} & \textbf{11.14} & \textbf{20.08}\\
\midrule
\midrule
EgoVLP & Zero-Shot & \underline{10.73} & \underline{24.70} & \underline{7.38} & \underline{16.53} \\
\bf Ours (Omnivore) & Zero-Shot & 9.29& 22.89 & 6.24 & 15.05 \\
\bf Ours (EgoVLP)& Zero-Shot & \textbf{11.57} & \textbf{27.41} & \textbf{7.87} & \textbf{18.70} \\
\bottomrule
\end{tabularx}
\vspace{-3mm}
\caption{\textbf{Step-Grounding on Ego4D Goal-Step~\cite{song2024ego4d}}. In the \emph{Supervised} setting, we compare different features extractors, including \ours, using VSLNet~\cite{zhang2020span} as grounding head. In the \emph{Zero-Shot} setting, we adopt the clustering approach of \ours.}
\vspace{-5mm}
\label{tab:goalstep_main}
\end{table}

\begin{table}[tb]
    \centering
    \scriptsize
    \setlength{\tabcolsep}{4.5pt}
    \begin{tabularx}{1.0\columnwidth}{@{}Xccccccc@{}}
        \toprule
        \multirow{2}{*}{\textbf{Method}} & \multirow{2}{*}{\textbf{Approach}} & \multicolumn{6}{c}{\textbf{mAP @ IoU}}                                 \\
                                         &                                    & 0.1                           & 0.2  & 0.3 & 0.4 & 0.5 & Avg  \\
        \midrule
        Omnivore~\cite{song2024ego4d}    & Supervised                         & $-$                                    & $-$  & $-$ & $-$ & $-$ & 10.3 \\
        EgoOnly~\cite{song2024ego4d}     & Supervised                         & $-$                                    & $-$  & $-$ & $-$ & $-$ & \textbf{13.6} \\
        \midrule
        EgoVLP & Supervised & 13.3 & 12.3 & 11.2 & 10.1 & 8.7 & 11.1 \\
        \textbf{Ours (EgoVLP)} & Supervised & 14.2 & 13.2 & 12.2 & 10.9 & 9.6 & \underline{12.0} \\
        \midrule
        \midrule
        EgoVLP & Zero-Shot & 11.8 & 9.7 & 8.3 & 6.7 & 5.1 & 8.3 \\
        \textbf{Ours (EgoVLP)} & Zero-Shot & 12.0 & 10.0 & 8.8  & 7.3 & 5.6 & 8.7 \\
        \bottomrule
    \end{tabularx}
    \vspace{-3mm}
    \caption{\textbf{Step Localization on Ego4D Goal-Step~\cite{song2024ego4d}}, in \emph{supervised} and \emph{zero-shot} settings. In the \emph{supervised} setting we use ActionFormer~\cite{zhang2022actionformer} as localization head, while \ours detects steps directly on the output features of the decoder using zero-shot matching with the steps taxonomy.
    }
    \vspace{-.5cm}
    \label{tab:goalstep_segm}
\end{table}
We evaluate \ours on the Step Grounding and Step Localization tasks from Ego4D Goal-Step~\cite{song2024ego4d}, demonstrating its ability to localize and classify procedural steps.

\smallskip
\noindent\textbf{Step Grounding.} 
This task aims to localize a procedure step given its description in natural language. Performance is measured with Recall at different IoU thresholds, as for EgoNLQ.
The supervised baseline proposed in~\cite{song2024ego4d} leverages VSLNet~\cite{zhang2020span} as grounding head on top of the Omnivore pre-extracted features. 
Instead, we adapt \ours to this task by clustering the video segments and selecting as prediction candidates the segment whose average visual features are most similar to the textual features of the query step. 
This allows to address the grounding task in \emph{zero-shot} without any additional training. 
We also evaluate the performance of \ours when used as a feature extractor in combination with VSLNet.
Table~\ref{tab:goalstep_main} shows that \ours consistently outperforms the Omnivore and EgoVLP baselines in the supervised setting. 
In \emph{zero-shot}, \ours beats the supervised counterpart on Top-5 Recall and achieves results close to the SOTA on the other metrics.

\smallskip
\noindent\textbf{Step Localization.}
This task aims to predict triplets \texttt{(start time, end time, label)} for all the procedure steps and substeps in the video. 
Similar to Step Grounding and EgoProceL, we adapt \ours to this task by clustering the output features to localize the steps and use the similarity between the visual and the textual features of the steps taxonomy to predict their labels (Table~\ref{tab:goalstep_segm}). 
We compare our approach with the two official baselines from Goal-Step~\cite{song2024ego4d}, which train an ActionFormer~\cite{zhang2022actionformer} localization head on top of  Omnivore~\cite{omnivore} and EgoOnly~\cite{wang2023ego} features.
Performance is evaluated in terms of mAP at different IoU thresholds. 
Notably, unlike other approaches that generate per-segment predictions and apply Soft-NMS~\cite{bodla2017soft} to filter overlaps, \ours produces non-overlapping step candidates and does not require any post-processing.
Remarkably, the zero-shot results of \ours demonstrate that our clustering approach effectively identifies action clusters, which are well aligned with the steps taxonomy.
Compared to supervised approaches that learn a direct mapping between the video input and the procedure steps, we argue that the steps detected by \ours emerge as composition of low-level patterns, as segments that represent similar patterns are clustered together.

\subsection{Ablation on the \ours components}
\begin{table}[t]
    \centering
    \scriptsize
    \setlength{\tabcolsep}{3.75pt}
    \begin{tabularx}{\columnwidth}{@{}>{\centering\arraybackslash}X >{\centering\arraybackslash}X >{\centering\arraybackslash}ccccccc@{}}
        \toprule
        \multirow{2}{*}{\shortstack[c]{\textbf{Align}\\[-0.24em]\textbf{Loss}}}
 & \multirow{2}{*}{\shortstack{\textbf{Func. Th.}\\\textbf{Cluster}}} & \multirow{2}{*}{\shortstack{\textbf{Func Th.}\\\textbf{Loss}}} & \multicolumn{2}{c}{\bf EgoMCQ} & \multicolumn{2}{c}{\bf EgoProceL} & \multicolumn{2}{c}{\bf Step-Grounding} \\
        & & & Inter & Intra & F1 & IoU & R@1 & R@5 \\
        \midrule
        \xmark & \xmark & \xmark & 90.6 & 57.2 & 40.0 & 21.9 & 10.73 & 24.70 \\
        \cmark & \xmark & \xmark & \textbf{91.8} & 59.5 & 43.8 & 24.1 & 11.27 & 27.35\\
        \cmark & \cmark & \xmark & \textbf{91.8} & \textbf{59.6} & 43.3 & 24.2 & 11.44 & 27.12 \\
        \cmark & \cmark & \cmark & 91.6 & \textbf{59.6} & \textbf{44.5} & \textbf{25.3} & \textbf{11.57} & \textbf{27.41} \\
        \bottomrule
    \end{tabularx}
    \vspace{-\baselineskip}
    \caption{\textbf{Ablation of the different components of \ours} on EgoMCQ, EgoProceL and Goal-Step, using EgoVLP features.}
    \vspace{-\baselineskip}
    \label{tab:ablation}
\end{table}

We analyze in Table~\ref{tab:ablation} the impact of the different components of \ours, using the EgoVLP backbone, on three significant tasks, namely EgoMCQ~\cite{lin2022egocentric}, Procedure Learning on EgoProceL~\cite{bansal2022my} and Step Grounding on Goal-Step~\cite{song2024ego4d}.
Compared to the baseline, the alignment loss $\mathcal{L}_{vna}$ significantly improves performance on all the tasks, demonstrating that the context-aware features of \ours effectively support various understanding tasks, particularly procedural ones. 
Training-time threads clustering has a more mild impact. 
However, the introduction of the \emph{functional threads} loss $\mathcal{L}_{ft}$ effectively guides the clustering process by encouraging samples within the same cluster to be closer in feature space, leading to better performance.

\subsection{Qualitative results}
We show in Fig.~\ref{fig:zs_segm} some success and failure cases of \ours in the \emph{zero-shot} Step Localization task on  Goal-Step. We observe that many failure cases of our approach are related to the ambiguous granularity of the step labels in the ground truth, which leads to confusion between steps that could be either steps or sub-steps, \eg, \emph{Cook or prepare the vegetables} and \emph{Cut the pepper} in Fig.~\ref{fig:zs_segm_a}. 
We provide additional qualitative results and a discussion on the emergence of procedure steps in the appendix.

\section{Conclusions}\label{sec:conclusions}
In this paper, we discuss the relevance of learning about the hierarchical structure of human behavior collected in egocentric videos. We propose \ours, a weakly-supervised method able to fully exploit functional threads to enhance reasoning capabilities for multiple downstream tasks. \ours delivers state of the art performance in zero-shot for procedural learning tasks, proving the effectiveness and importance of using functional reasoning at multiple levels. \ours features proved their suitability for video-text alignment tasks, outperforming foundational models features.

\section*{Acknowledgments}
This study was carried out within the FAIR - Future Artificial Intelligence Research and received funding from the European Union Next-GenerationEU (PIANO NAZIONALE DI RIPRESA E RESILIENZA (PNRR) – MISSIONE 4 COMPONENTE 2, INVESTIMENTO 1.3 – D.D. 1555 11/10/2022, PE00000013). This manuscript reflects only the authors’ views and opinions, neither the European Union nor the European Commission can be considered responsible for them. We acknowledge the CINECA award under the ISCRA initiative, for the availability of high performance computing resources and support. 

{
    \small
    \bibliographystyle{ieeenat_fullname}
    \bibliography{main}

\begin{thebibliography}{61}
\providecommand{\natexlab}[1]{#1}
\providecommand{\url}[1]{\texttt{#1}}
\expandafter\ifx\csname urlstyle\endcsname\relax
  \providecommand{\doi}[1]{doi: #1}\else
  \providecommand{\doi}{doi: \begingroup \urlstyle{rm}\Url}\fi

\bibitem[Afouras et~al.(2023)Afouras, Mavroudi, Nagarajan, Wang, and Torresani]{afouras2023ht}
Triantafyllos Afouras, Effrosyni Mavroudi, Tushar Nagarajan, Huiyu Wang, and Lorenzo Torresani.
\newblock Ht-step: Aligning instructional articles with how-to videos.
\newblock In \emph{NeurIPS}, 2023.

\bibitem[Ashutosh et~al.(2023{\natexlab{a}})Ashutosh, Girdhar, Torresani, and Grauman]{hiervl}
Kumar Ashutosh, Rohit Girdhar, Lorenzo Torresani, and Kristen Grauman.
\newblock Hiervl: Learning hierarchical video-language embeddings.
\newblock In \emph{CVPR}, 2023{\natexlab{a}}.

\bibitem[Ashutosh et~al.(2023{\natexlab{b}})Ashutosh, Ramakrishnan, Afouras, and Grauman]{ashutosh2023video}
Kumar Ashutosh, Santhosh~Kumar Ramakrishnan, Triantafyllos Afouras, and Kristen Grauman.
\newblock Video-mined task graphs for keystep recognition in instructional videos.
\newblock \emph{NeurIPS}, 2023{\natexlab{b}}.

\bibitem[Bansal et~al.(2022)Bansal, Arora, and Jawahar]{bansal2022my}
Siddhant Bansal, Chetan Arora, and CV Jawahar.
\newblock My view is the best view: Procedure learning from egocentric videos.
\newblock In \emph{ECCV}, 2022.

\bibitem[Bansal et~al.(2024)Bansal, Arora, and Jawahar]{bansal2024united}
Siddhant Bansal, Chetan Arora, and CV Jawahar.
\newblock United we stand, divided we fall: Unitygraph for unsupervised procedure learning from videos.
\newblock In \emph{WACV}, 2024.

\bibitem[Bodla et~al.(2017)Bodla, Singh, Chellappa, and Davis]{bodla2017soft}
Navaneeth Bodla, Bharat Singh, Rama Chellappa, and Larry~S Davis.
\newblock Soft-nms--improving object detection with one line of code.
\newblock In \emph{ICCV}, 2017.

\bibitem[Botvinick et~al.(2009)Botvinick, Niv, and Barto]{botvinick2009hierarchically}
Matthew~M Botvinick, Yael Niv, and Andew~G Barto.
\newblock Hierarchically organized behavior and its neural foundations: A reinforcement learning perspective.
\newblock \emph{Cognition}, 2009.

\bibitem[Chowdhury et~al.(2024)Chowdhury, Chandra, and Roy]{chowdhuryopel}
Sayeed~Shafayet Chowdhury, Soumyadeep Chandra, and Kaushik Roy.
\newblock Opel: Optimal transport guided procedure learning.
\newblock In \emph{NeurIPS}, 2024.

\bibitem[Cooper and Shallice(2006)]{cooper2006hierarchical}
Richard~P Cooper and Tim Shallice.
\newblock Hierarchical schemas and goals in the control of sequential behavior.
\newblock \emph{Psychological Review}, 2006.

\bibitem[De~la Torre et~al.(2008)De~la Torre, Hodgins, Montano, Valcarcel, Forcada, and Macey]{cmu_mmac}
Fernando De~la Torre, Jessica Hodgins, J Montano, S Valcarcel, R Forcada, and J Macey.
\newblock Carnegie mellon university multimodal activity (cmu-mmac) database, 2008.

\bibitem[Dvornik et~al.(2023)Dvornik, Hadji, Zhang, Derpanis, Wildes, and Jepson]{dvornik2023stepformer}
Nikita Dvornik, Isma Hadji, Ran Zhang, Konstantinos~G Derpanis, Richard~P Wildes, and Allan~D Jepson.
\newblock Stepformer: Self-supervised step discovery and localization in instructional videos.
\newblock In \emph{CVPR}, 2023.

\bibitem[Elhamifar and Huynh(2020)]{elhamifar2020self}
Ehsan Elhamifar and Dat Huynh.
\newblock Self-supervised multi-task procedure learning from instructional videos.
\newblock In \emph{ECCV}, 2020.

\bibitem[Feichtenhofer et~al.(2019)Feichtenhofer, Fan, Malik, and He]{slowfast}
Christoph Feichtenhofer, Haoqi Fan, Jitendra Malik, and Kaiming He.
\newblock Slowfast networks for video recognition.
\newblock In \emph{ICCV}, 2019.

\bibitem[Gao and Ji(2019)]{gao2019graph}
Hongyang Gao and Shuiwang Ji.
\newblock Graph u-nets.
\newblock In \emph{ICML}, 2019.

\bibitem[Girdhar et~al.(2022)Girdhar, Singh, Ravi, van~der Maaten, Joulin, and Misra]{omnivore}
Rohit Girdhar, Mannat Singh, Nikhila Ravi, Laurens van~der Maaten, Armand Joulin, and Ishan Misra.
\newblock Omnivore: A single model for many visual modalities.
\newblock In \emph{CVPR}, 2022.

\bibitem[Goletto et~al.(2024)Goletto, Nagarajan, Averta, and Damen]{goletto2025amego}
Gabriele Goletto, Tushar Nagarajan, Giuseppe Averta, and Dima Damen.
\newblock Amego: Active memory from long egocentric videos.
\newblock In \emph{ECCV}, 2024.

\bibitem[Grauman et~al.(2022)Grauman, Westbury, Byrne, Chavis, Furnari, Girdhar, Hamburger, Jiang, Liu, Liu, et~al.]{ego4d}
Kristen Grauman, Andrew Westbury, Eugene Byrne, Zachary Chavis, Antonino Furnari, Rohit Girdhar, Jackson Hamburger, Hao Jiang, Miao Liu, Xingyu Liu, et~al.
\newblock Ego4d: Around the world in 3,000 hours of egocentric video.
\newblock In \emph{CVPR}, 2022.

\bibitem[Hu et~al.(2022)Hu, yelong shen, Wallis, Allen-Zhu, Li, Wang, Wang, and Chen]{hu2021lora}
Edward~J Hu, yelong shen, Phillip Wallis, Zeyuan Allen-Zhu, Yuanzhi Li, Shean Wang, Lu Wang, and Weizhu Chen.
\newblock Lo{RA}: Low-rank adaptation of large language models.
\newblock In \emph{ICLR}, 2022.

\bibitem[Huang et~al.(2020)Huang, Sugano, and Sato]{huang2020improving}
Yifei Huang, Yusuke Sugano, and Yoichi Sato.
\newblock Improving action segmentation via graph-based temporal reasoning.
\newblock In \emph{CVPR}, 2020.

\bibitem[Islam et~al.(2024)Islam, Ho, Yang, Nagarajan, Torresani, and Bertasius]{islam2024video}
Md~Mohaiminul Islam, Ngan Ho, Xitong Yang, Tushar Nagarajan, Lorenzo Torresani, and Gedas Bertasius.
\newblock Video recap: Recursive captioning of hour-long videos.
\newblock In \emph{CVPR}, 2024.

\bibitem[Jang et~al.(2019)Jang, Sullivan, Ludwig, Gilchrist, Damen, and Mayol-Cuevas]{epic_tent}
Youngkyoon Jang, Brian Sullivan, Casimir Ludwig, Iain Gilchrist, Dima Damen, and Walterio Mayol-Cuevas.
\newblock Epic-tent: An egocentric video dataset for camping tent assembly.
\newblock In \emph{ICCVW}, 2019.

\bibitem[Jia et~al.(2022)Jia, Lei, Zhu, and Huang]{jia2022egotaskqa}
Baoxiong Jia, Ting Lei, Song-Chun Zhu, and Siyuan Huang.
\newblock Egotaskqa: Understanding human tasks in egocentric videos.
\newblock In \emph{NeurIPS}, 2022.

\bibitem[Kowal et~al.(2024)Kowal, Siam, Islam, Bruce, Wildes, and Derpanis]{kowal2024quantifying}
Matthew Kowal, Mennatullah Siam, Md~Amirul Islam, Neil~DB Bruce, Richard~P Wildes, and Konstantinos~G Derpanis.
\newblock Quantifying and learning static vs. dynamic information in deep spatiotemporal networks.
\newblock \emph{IEEE TPAMI}, 2024.

\bibitem[Kuehne et~al.(2014)Kuehne, Arslan, and Serre]{Kuehne12}
H. Kuehne, A.~B. Arslan, and T. Serre.
\newblock The language of actions: Recovering the syntax and semantics of goal-directed human activities.
\newblock In \emph{CVPR}, 2014.

\bibitem[Kuhn(1955)]{kuhn1955hungarian}
Harold~W Kuhn.
\newblock The hungarian method for the assignment problem.
\newblock \emph{Naval research logistics quarterly}, 1955.

\bibitem[Kumar et~al.(2022)Kumar, Haresh, Ahmed, Konin, Zia, and Tran]{kumar2022unsupervised}
Sateesh Kumar, Sanjay Haresh, Awais Ahmed, Andrey Konin, M~Zeeshan Zia, and Quoc-Huy Tran.
\newblock Unsupervised action segmentation by joint representation learning and online clustering.
\newblock In \emph{CVPR}, 2022.

\bibitem[Li and Todorovic(2020)]{li2020set}
Jun Li and Sinisa Todorovic.
\newblock Set-constrained viterbi for set-supervised action segmentation.
\newblock In \emph{CVPR}, 2020.

\bibitem[Li et~al.(2018)Li, Liu, and Rehg]{li2018eye}
Yin Li, Miao Liu, and James~M Rehg.
\newblock In the eye of beholder: Joint learning of gaze and actions in first person video.
\newblock In \emph{ECCV}, 2018.

\bibitem[Lin et~al.(2022{\natexlab{a}})Lin, Wang, Soldan, Wray, Yan, XU, Gao, Tu, Zhao, Kong, et~al.]{lin2022egocentric}
Kevin~Qinghong Lin, Jinpeng Wang, Mattia Soldan, Michael Wray, Rui Yan, Eric~Z XU, Difei Gao, Rong-Cheng Tu, Wenzhe Zhao, Weijie Kong, et~al.
\newblock Egocentric video-language pretraining.
\newblock In \emph{NeurIPS}, 2022{\natexlab{a}}.

\bibitem[Lin et~al.(2022{\natexlab{b}})Lin, Petroni, Bertasius, Rohrbach, Chang, and Torresani]{lin2022learning}
Xudong Lin, Fabio Petroni, Gedas Bertasius, Marcus Rohrbach, Shih-Fu Chang, and Lorenzo Torresani.
\newblock Learning to recognize procedural activities with distant supervision.
\newblock In \emph{CVPR}, 2022{\natexlab{b}}.

\bibitem[Mangalam et~al.(2023)Mangalam, Akshulakov, and Malik]{mangalam2023egoschema}
Karttikeya Mangalam, Raiymbek Akshulakov, and Jitendra Malik.
\newblock Egoschema: A diagnostic benchmark for very long-form video language understanding.
\newblock In \emph{NeurIPS}, 2023.

\bibitem[Mavroudi et~al.(2023)Mavroudi, Afouras, and Torresani]{mavroudi2023learning}
Effrosyni Mavroudi, Triantafyllos Afouras, and Lorenzo Torresani.
\newblock Learning to ground instructional articles in videos through narrations.
\newblock In \emph{ICCV}, 2023.

\bibitem[Melas-Kyriazi et~al.(2022)Melas-Kyriazi, Rupprecht, Laina, and Vedaldi]{melas2022deep}
Luke Melas-Kyriazi, Christian Rupprecht, Iro Laina, and Andrea Vedaldi.
\newblock Deep spectral methods: A surprisingly strong baseline for unsupervised semantic segmentation and localization.
\newblock In \emph{CVPR}, 2022.

\bibitem[Nagarajan et~al.(2019)Nagarajan, Feichtenhofer, and Grauman]{nagarajan2019grounded}
Tushar Nagarajan, Christoph Feichtenhofer, and Kristen Grauman.
\newblock Grounded human-object interaction hotspots from video.
\newblock In \emph{ICCV}, 2019.

\bibitem[Nagarajan et~al.(2020)Nagarajan, Li, Feichtenhofer, and Grauman]{nagarajan2020ego}
Tushar Nagarajan, Yanghao Li, Christoph Feichtenhofer, and Kristen Grauman.
\newblock Ego-topo: Environment affordances from egocentric video.
\newblock In \emph{CVPR}, 2020.

\bibitem[Naing and Elhamifar(2020)]{naing2020procedure}
Zwe Naing and Ehsan Elhamifar.
\newblock Procedure completion by learning from partial summaries.
\newblock In \emph{BMVC}, 2020.

\bibitem[Peirone et~al.(2024)Peirone, Pistilli, Alliegro, and Averta]{peirone2024backpack}
Simone~Alberto Peirone, Francesca Pistilli, Antonio Alliegro, and Giuseppe Averta.
\newblock A backpack full of skills: Egocentric video understanding with diverse task perspectives.
\newblock In \emph{CVPR}, 2024.

\bibitem[Peirone et~al.(2025)Peirone, Pistilli, Alliegro, Tommasi, and Averta]{peirone2025hier}
Simone~Alberto Peirone, Francesca Pistilli, Antonio Alliegro, Tatiana Tommasi, and Giuseppe Averta.
\newblock Hier-egopack: Hierarchical egocentric video understanding with diverse task perspectives.
\newblock \emph{arXiv preprint arXiv:2502.02487}, 2025.

\bibitem[Pramanick et~al.(2023)Pramanick, Song, Nag, Lin, Shah, Shou, Chellappa, and Zhang]{pramanick2023egovlpv2}
Shraman Pramanick, Yale Song, Sayan Nag, Kevin~Qinghong Lin, Hardik Shah, Mike~Zheng Shou, Rama Chellappa, and Pengchuan Zhang.
\newblock Egovlpv2: Egocentric video-language pre-training with fusion in the backbone.
\newblock In \emph{ICCV}, 2023.

\bibitem[Price et~al.(2022)Price, Vondrick, and Damen]{price2022unweavenet}
Will Price, Carl Vondrick, and Dima Damen.
\newblock Unweavenet: Unweaving activity stories.
\newblock In \emph{CVPR}, 2022.

\bibitem[Ragusa et~al.(2023)Ragusa, Furnari, and Farinella]{ragusa_MECCANO_2023}
Francesco Ragusa, Antonino Furnari, and Giovanni~Maria Farinella.
\newblock Meccano: A multimodal egocentric dataset for humans behavior understanding in the industrial-like domain.
\newblock \emph{CVIU}, 2023.

\bibitem[Richard et~al.(2018)Richard, Kuehne, Iqbal, and Gall]{richard2018neuralnetwork}
Alexander Richard, Hilde Kuehne, Ahsan Iqbal, and Juergen Gall.
\newblock Neuralnetwork-viterbi: A framework for weakly supervised video learning.
\newblock In \emph{CVPR}, 2018.

\bibitem[Sanh(2019)]{sanh2019distilbert}
V Sanh.
\newblock Distilbert, a distilled version of bert: smaller, faster, cheaper and lighter.
\newblock \emph{arXiv preprint arXiv:1910.01108}, 2019.

\bibitem[Sarfraz et~al.(2021)Sarfraz, Murray, Sharma, Diba, Van~Gool, and Stiefelhagen]{sarfraz2021temporally}
Saquib Sarfraz, Naila Murray, Vivek Sharma, Ali Diba, Luc Van~Gool, and Rainer Stiefelhagen.
\newblock Temporally-weighted hierarchical clustering for unsupervised action segmentation.
\newblock In \emph{CVPR}, 2021.

\bibitem[Seminara et~al.(2024)Seminara, Farinella, and Furnari]{seminara2024differentiable}
Luigi Seminara, Giovanni~Maria Farinella, and Antonino Furnari.
\newblock Differentiable task graph learning: Procedural activity representation and online mistake detection from egocentric videos.
\newblock In \emph{NeurIPS}, 2024.

\bibitem[Shen and Elhamifar(2024)]{shen2024progress}
Yuhan Shen and Ehsan Elhamifar.
\newblock Progress-aware online action segmentation for egocentric procedural task videos.
\newblock In \emph{CVPR}, 2024.

\bibitem[Shi and Malik(2000)]{shi2000normalized}
Jianbo Shi and Jitendra Malik.
\newblock Normalized cuts and image segmentation.
\newblock \emph{IEEE TPAMI}, 2000.

\bibitem[Song et~al.(2024)Song, Byrne, Nagarajan, Wang, Martin, and Torresani]{song2024ego4d}
Yale Song, Eugene Byrne, Tushar Nagarajan, Huiyu Wang, Miguel Martin, and Lorenzo Torresani.
\newblock Ego4d goal-step: Toward hierarchical understanding of procedural activities.
\newblock In \emph{NeurIPS}, 2024.

\bibitem[Stein and McKenna(2013)]{stein2013combining}
Sebastian Stein and Stephen~J McKenna.
\newblock Combining embedded accelerometers with computer vision for recognizing food preparation activities.
\newblock In \emph{Proceedings of the 2013 ACM international joint conference on Pervasive and ubiquitous computing}, 2013.

\bibitem[Tong et~al.(2022)Tong, Song, Wang, and Wang]{tong2022videomae}
Zhan Tong, Yibing Song, Jue Wang, and Limin Wang.
\newblock Videomae: Masked autoencoders are data-efficient learners for self-supervised video pre-training.
\newblock In \emph{NeurIPS}, 2022.

\bibitem[Von~Luxburg(2007)]{von2007tutorial}
Ulrike Von~Luxburg.
\newblock A tutorial on spectral clustering.
\newblock \emph{Statistics and computing}, 17:\penalty0 395--416, 2007.

\bibitem[Wang et~al.(2023{\natexlab{a}})Wang, Singh, and Torresani]{wang2023ego}
Huiyu Wang, Mitesh~Kumar Singh, and Lorenzo Torresani.
\newblock Ego-only: Egocentric action detection without exocentric transferring.
\newblock In \emph{ICCV}, 2023{\natexlab{a}}.

\bibitem[Wang et~al.(2024)Wang, Misra, Zeng, Girdhar, and Darrell]{wang2024videocutler}
Xudong Wang, Ishan Misra, Ziyun Zeng, Rohit Girdhar, and Trevor Darrell.
\newblock Videocutler: Surprisingly simple unsupervised video instance segmentation.
\newblock In \emph{CVPR}, 2024.

\bibitem[Wang et~al.(2023{\natexlab{b}})Wang, Shen, Yuan, Du, Li, Hu, Crowley, and Vaufreydaz]{wang2023tokencut}
Yangtao Wang, Xi Shen, Yuan Yuan, Yuming Du, Maomao Li, Shell~Xu Hu, James~L Crowley, and Dominique Vaufreydaz.
\newblock Tokencut: Segmenting objects in images and videos with self-supervised transformer and normalized cut.
\newblock \emph{IEEE TPAMI}, 2023{\natexlab{b}}.

\bibitem[Zhang et~al.(2022)Zhang, Wu, and Li]{zhang2022actionformer}
Chen-Lin Zhang, Jianxin Wu, and Yin Li.
\newblock Actionformer: Localizing moments of actions with transformers.
\newblock In \emph{ECCV}, 2022.

\bibitem[Zhang et~al.(2020)Zhang, Sun, Jing, and Zhou]{zhang2020span}
Hao Zhang, Aixin Sun, Wei Jing, and Joey~Tianyi Zhou.
\newblock Span-based localizing network for natural language video localization.
\newblock In \emph{Proceedings of the 58th Annual Meeting of the Association for Computational Linguistics}, 2020.

\bibitem[Zhao et~al.(2023)Zhao, Misra, Kr{\"a}henb{\"u}hl, and Girdhar]{zhao2023learning}
Yue Zhao, Ishan Misra, Philipp Kr{\"a}henb{\"u}hl, and Rohit Girdhar.
\newblock Learning video representations from large language models.
\newblock In \emph{CVPR}, 2023.

\bibitem[Zhong et~al.(2023)Zhong, Yu, Bai, Li, Yan, and Li]{zhong2023learning}
Yiwu Zhong, Licheng Yu, Yang Bai, Shangwen Li, Xueting Yan, and Yin Li.
\newblock Learning procedure-aware video representation from instructional videos and their narrations.
\newblock In \emph{CVPR}, 2023.

\bibitem[Zhou et~al.(2023)Zhou, Mart{\'\i}n-Mart{\'\i}n, Kapadia, Savarese, and Niebles]{zhou2023procedure}
Honglu Zhou, Roberto Mart{\'\i}n-Mart{\'\i}n, Mubbasir Kapadia, Silvio Savarese, and Juan~Carlos Niebles.
\newblock Procedure-aware pretraining for instructional video understanding.
\newblock In \emph{CVPR}, 2023.

\bibitem[Zhou et~al.(2018)Zhou, Xu, and Corso]{zhou2018towards}
Luowei Zhou, Chenliang Xu, and Jason Corso.
\newblock Towards automatic learning of procedures from web instructional videos.
\newblock In \emph{AAAI}, 2018.

\bibitem[Zhukov et~al.(2019)Zhukov, Alayrac, Cinbis, Fouhey, Laptev, and Sivic]{zhukov2019cross}
Dimitri Zhukov, Jean-Baptiste Alayrac, Ramazan~Gokberk Cinbis, David Fouhey, Ivan Laptev, and Josef Sivic.
\newblock Cross-task weakly supervised learning from instructional videos.
\newblock In \emph{CVPR}, 2019.

\end{thebibliography}
}

\clearpage
\maketitlesupplementary

Sec.~\ref{sec:supp_datasets_tasks} provides further details on the datasets and tasks used in this work. 
Sec.~\ref{sec:supp_impl_details} presents additional implementation details and a discussion of some key design choices behind \ours.
Sec.~\ref{sec:supp_exps} evaluates different clustering algorithms for \ours on the Step Grounding and Procedure Learning tasks.
Sec.~\ref{sec:interpretability} discusses the unsupervised emergence of procedural steps in \ours's features space.
Sec.~\ref{sec:opel_omnivore} analyzes the impact of using a more informative backbone in the previous SOTA on the EgoProceL benchmark.
Finally, Sec.~\ref{sec:qualitatives} presents additional qualitative results on the Step Localization task.

\setcounter{section}{0}
\renewcommand{\thesection}{\Alph{section}}

\section{Dataset and task details}\label{sec:supp_datasets_tasks}

\subsection{Ego4D}
Ego4D is a large scale egocentric vision dataset with $3670$ hours of daily-life activities captured from $931$ subjects around the world. Videos are annotated with fine-grained textual descriptions of the activities performed by the camera wearer or other participants in the scene, \eg, \emph{``\#C C stirs food in a frying pan with a spoon in his right hand''}, and with task-specific annotations on a subset of the videos for a wide range of tasks, including episodic memory, spatial and temporal grounding of the interactions, forecasting, etc.
We focus our analysis on two benchmark, namely EgoMCQ and EgoNLQ.

\paragraph{EgoMCQ.} EgoMCQ is a development benchmark introduced with EgoVLP~\cite{lin2022egocentric} to validate the quality of video-language pretraining models. It features $39k$ multiple-choice questions generated from Ego4D annotations. Given a textual query and five candidate video clips, the task is to identify the correct clip. Candidates may belong to the same video (\emph{intra-video}) or from different videos (\emph{inter-video}). Performance is evaluated in terms of accuracy.

\paragraph{EgoNLQ.} EgoNLQ is a temporal grounding task that requires multi-modal video and language reasoning. Given a textual query from a set of predefined templates, the goal is to identify the temporal boundaries (start and end timestamps) of the video segment that answers the query. The benchmark includes $13.6k$ / $4.5k$ / $4.4k$ queries in the train, validation and test splits respectively. We follow previous works in video-language pre-training~\cite{lin2022egocentric,pramanick2023egovlpv2,zhao2023learning} and evaluate \ours on this task using VSLNet~\cite{zhang2020span} as grounding head, using the same hyper-parameter tuning recipe as EgoVLP~\cite{lin2022egocentric} and reporting results on the validation set.
As for EgoNLQ, performance is evaluated in terms of Top-1 and Top-5 Recall at different Intersection over Union ($0.3$ and $0.5$) between the predicted and the ground truth segments.

\subsection{Goal-Step} Goal-Step~\cite{song2024ego4d} extends the Ego4D dataset with annotations of hierarchical activity labels, identifying goals, steps and substeps in procedural activities. It provides dense annotations for $48k$ procedural step segments ($480$ hours), from a taxonomy of $501$ labels. 
We evaluate \ours on the Step Grounding and Step Localization tasks.

\paragraph{Step Grounding.} Step Grounding is a temporal grounding task, in which the goal is to recognize the temporal boundaries of a procedural step given its description in natural language. 
For supervised experiments we use the same architecture of the baseline (VSLNet~\cite{zhang2020span}) with the same hyper-parameters and report performance as the average of 8 runs. When using EgoVLP features we extend the number of samples in the input sequence from $128$ to $256$.
Performance is evaluated in terms of Top-1 and Top-5 Recall at different Intersection over Union ($0.3$ and $0.5$) between the predicted and the ground truth segments.

\paragraph{Step Localization.} Step Localization is more closely related to action segmentation. Given a long video, the goal is to find all the procedure steps in the video with their corresponding start/end time and label according to the Goal-Step taxonomy.
Models are trained and evaluated on steps and substeps without distinctions.
The supervised models use ActionFormer~\cite{zhang2022actionformer} as localization head, with base learning rate of 2e-4 and training for 32 epochs with linear warm-up for 16 epochs.
Performance is evaluated in terms of mAP at different Intersection over Union (IoU) thresholds between the predicted and the ground truth segments.

\subsection{EgoProceL}
EgoProceL~\cite{bansal2022my} collects multiple egocentric vision datasets focusing on procedural tasks that require multiple steps, \eg, \emph{Preparing a salad} or \emph{Assemblying a PC}: MECCANO~\cite{ragusa_MECCANO_2023}, Epic-Tents~\cite{epic_tent}, CMU-MMAC~\cite{cmu_mmac}, EGTEA~\cite{li2018eye} and PC Assembly/Disassembly~\cite{bansal2022my}.
Table~\ref{tab:supp_egoprocel_stats} reports the number of videos and key-steps in each task of the dataset. 
Annotations assign each video frame to a specific key-step of the corresponding task.
\begin{table}[t]
    \centering
    \scriptsize
    \begin{tabularx}{\columnwidth}{@{}Xcc@{}}
        \toprule
        \textbf{Task} & \textbf{Videos Count} & \textbf{Key-steps Count}\\
        \midrule
        PC Assembly~\cite{bansal2022my} & 14 & 9\\
        PC Disassembly~\cite{bansal2022my} & 15 & 9\\
        MECCANO~\cite{ragusa_MECCANO_2023} & 17 & 17\\
        Epic-Tents~\cite{epic_tent} & 29 & 12\\
        \midrule
        \midrule
        \multicolumn{3}{c}{CMU-MMAC~\cite{cmu_mmac}} \\
        Brownie & 34 & 9\\
        Eggs & 33 & 8\\
        Pepperoni Pizza & 33 & 5\\
        Salad & 34 & 9\\
        Sandwich & 31 & 4\\
        \midrule
        \midrule
        \multicolumn{3}{c}{EGTEA+~\cite{li2018eye}}\\
        Bacon and Eggs & 16 & 11\\
        Cheese Burger & 10 & 10\\
        Continental Breakfast & 12 & 10\\
        Greek Salad & 10 & 4\\
        Pasta Salad & 19 & 8\\
        Hot Box Pizza & 6 & 8\\
        Turkey Sandwich & 13 & 6\\
        \bottomrule
    \end{tabularx}
    \vspace{-2.5mm}
    \caption{\textbf{Number of videos and key-steps in EgoProceL~\cite{bansal2022my}.}}
    \label{tab:supp_egoprocel_stats}
\end{table}
We evaluate \ours on the \textbf{Procedure Learning} task, following the same evaluation protocol of previous works~\cite{bansal2022my,bansal2024united,chowdhuryopel}. 
Specifically, we compute framewise step assignment and evaluate the F1-score and Intersection over Union (IoU) between the predicted steps and the ground truth labels for each step separately. 
The F1-score is computed as the harmonic mean of precision and recall. Precision is the proportion of correctly identified key-step frames out of all frames predicted to be key-steps, while recall is the proportion of correctly identified key-step frames out of the total number of actual key-step frames.
Predictions and ground truth labels are matched using the Hungarian algorithm, following previous works~\cite{bansal2022my,chowdhuryopel}.

\section{Additional implementation details}\label{sec:supp_impl_details}
\ours follows an encoder-decoder architecture with three stages, each comprising three layers of TDGC~\cite{peirone2025hier}, with hidden feature size $768$ and the threshold for temporal graph connectivity $d$ is set to $1$. 
Input features are first projected to size $768$ using a linear layer.
For $\mathcal{L}_{vna}$ and $\mathcal{L}_{ft}$, we set the temperature parameter to $\tau = 0.05$.
When evaluating \ours on EgoMCQ, we assume that only a single functional thread is present in the input video, given the short duration of the clip, and disable the functional threads clustering of the decoder.

\paragraph{Text-encoder fine-tuning.}\label{supp:text-finetuning}
\begin{table}[h]
    \centering
    \scriptsize
    \begin{tabular}{@{}lccc@{}}
        \toprule
         \bf \multirow{2}{*}{Strategy} & \bf \multirow{2}{*}{\shortstack{Trainable\\Params}} & \multicolumn{2}{c}{\bf EgoMCQ} \\
         & & \bf Inter & \bf Intra \\
         \midrule
         Frozen & 20.10 M & 84.2 & 46.0 \\
         \midrule
         LoRa~\cite{hu2021lora} & 20.99 M & \underline{88.2} & \underline{49.7} \\
         Full Fine-Tuning & 86.47 M & \bf 90.3 & \bf 53.3 \\
         \bottomrule
    \end{tabular}
    \vspace{-2.5mm}
    \caption{\textbf{Comparison of different fine-tuning strategies for the text-encoder of \ours}, using Omnivore features and measuring performance on EgoMCQ. Full fine-tuning significantly improves accuracy.}
    \label{tab:supp_text-finetuning}
\end{table}
EgoVLP~\cite{lin2022egocentric} and LaViLa~\cite{zhao2023learning} were trained for video-text alignment. Therefore, when building \ours on these backbones we reuse their respective text encoders, with no additional training.
Instead, Omnivore was not trained for video-text alignment and does not have a text encoder. In this case, we bootstrap the text encoder of \ours from a pretrained DistillBERT~\cite{sanh2019distilbert} and fine-tune it during the the training process. We experiment different strategies to fine-tune the text encoder, using LoRa~\cite{hu2021lora} to reduce the number of trainable parameters or fully updating the text encoder, as shown in Table~\ref{tab:supp_text-finetuning}.
While LoRa provides a significant improvement compared to the frozen text encoder, the gap with the full fine-tuning is consistent. 
Remarkably, with little computational overhead (training lasts less than 20 GPU hours), \ours reaches performance close to that of EgoVLP, despite not being trained end-to-end on Ego4D.

\paragraph{Impact of the context window in EgoMCQ.}\label{sec:supp_context_impact}
\begin{table}[h]
    \centering
    \scriptsize
    \setlength{\tabcolsep}{5pt}
    \begin{tabular}{@{}c|cc|cc|cc@{}}
        \toprule
         \multirow{2}{*}{$\Delta$} & \multicolumn{2}{c|}{\bf EgoVLP~\cite{lin2022egocentric}} & \multicolumn{2}{c|}{\bf LaViLa~\cite{zhao2023learning}} & \multicolumn{2}{c}{\bf \ours (EgoVLP)} \\
         & Inter & Intra & Inter & Intra & Inter & Intra \\
         \midrule
         N/A (paper) & 90.6 & 57.2 & 94.5 & 63.1 & $-$ & $-$ \\
         \midrule
         0 & 90.7 & \textbf{53.4} & 93.9 & \textbf{57.9} & 89.0 & 52.4 \\
         1 & \textbf{91.0} & 52.5 & \textbf{94.1} & 56.7 & 90.9 & 57.4 \\
         2 & 90.8 & 48.7 & 93.6 & 52.5 & 91.3 & 58.8 \\
         4 & 89.9 & 42.2 & 93.1 & 44.8 & \textbf{91.8} & \textbf{59.5} \\ 
         \bottomrule
    \end{tabular}
    \vspace{-2.5mm}
    \caption{\textbf{Impact of the additional context window on EgoMCQ Accuracy (\%).} The first row refers to the original results, as reported in their respective papers.}
    \label{tab:supp_impact_ctx_window}
\end{table}
\ours is built on dense pre-extracted features from fixed size segments (16 frames) of the video, using a pre-trained backbone, \eg, EgoVLP~\cite{lin2022egocentric} or LaViLa~\cite{zhao2023learning}.
Each segment is mapped to a node of the input graph $\mathcal{G}$.
We adapt the evaluation process for \ours to work with pre-extracted features. 
Specifically, when evaluating \ours on benchmarks that require a fixed size input, \eg, EgoMCQ, the nodes correspond to all video segments that fall between the start $t_s$ and end timestamps $t_e$ of the input.
Since clips in EgoMCQ are very short ($0.84s$ on average), we slightly extend the clip segment by a context window $\Delta$ to provide additional temporal context and ensure and the resulting graph has a reasonable number of nodes for processing.
We adapt EgoVLP and LaViLa to our setting, \ie, using dense features extracted from video segments with additional temporal context, and evaluate the impact of this additional temporal context on EgoVLP and LaViLa in Table~\ref{tab:supp_impact_ctx_window}, showing that this additional context does not trivially translate to better performance on this benchmark.
In contrast, \ours is trained to exploit such additional temporal context and achieves best performance when used in combination with a larger input window ($\Delta = 4$). At the same time, \ours is quite robust even to shorter context windows.

\paragraph{Video-Narrations alignment window.}\label{sec:supp_vna_window}
\begin{table}[ht]
    \centering
    \scriptsize
    \begin{tabular}{@{}cccc@{}}
        \toprule
         \bf \multirow{2}{*}{$\alpha$} & \bf \multirow{2}{*}{$\beta$} & \multicolumn{2}{c}{\bf EgoMCQ} \\
         & & Inter & Intra \\
         \midrule
         1 & \emph{all} & 91.8 & \textbf{59.5} \\
         1 & 4 & 91.8 & 57.4 \\
         1 & 16 & \textbf{92.0} & \underline{58.5} \\
         \midrule
         2 & \emph{all} & 91.5 & \textbf{59.5} \\
         2 & 4 & 91.5 & 56.5 \\
         2 & 16 & \underline{91.9} & 58.2 \\
         \bottomrule
    \end{tabular}
    \vspace{-2.5mm}
    \caption{\textbf{Ablation on the size of the video-narrations alignment window.} For $\beta$, \emph{all} means that all narrations from the same video that are not part of the positives set are considered as negatives.}
    \label{tab:supp_vna}
\end{table}
We evaluate in Table~\ref{tab:supp_vna}, different choices for the $\alpha$ and $\beta$ parameters that control the size of the alignment window in $\mathcal{L}_{vna}$.
$\alpha$ controls the window size for positive samples, with higher values resulting in narrower windows. $\beta$ controls the window for sampling negatives narrations from the same video. Higher values indicate larger windows, with \emph{all} meaning that all narrations from the videos are taken as negative, except the ones that fall inside the positives window.
The $\alpha$ parameter has little impact on both \emph{inter} and \emph{intra} accuracy.
The $\beta$ parameter has a more noticeable impact on performance, with best results when all intra-video narrations are used as negatives.

\subsection{Additional details on the Cut\&Match module}
The Cut\&Match module updates the connectivity of a video graph $\mathcal{G}$ in the \ours architecture to connect regions, \ie, video segments, that may be temporally distant but encode functionally related actions. This is achieved by grouping the graph nodes into $K$ different partitions based on features cosine similarity using spectral clustering.
As a result, the input graph $\mathcal{G}$ is partitioned into $K$ sub-graphs $\{\tilde{\mathcal{G}}^{l+1}_{d,1}, \dots, \tilde{\mathcal{G}}^{l+1}_{d,K}\}$. Temporal reasoning is implemented on each sub-graph separately and nodes are then mapped back to the original graph.

\smallskip
\noindent\textbf{Approximated graph partitioning.} To efficiently implement the graph partitioning step on a batch of graphs, we approximate node partitioning by uniformly sub-sampling each graph to a fixed number of nodes based on the node timestamps.
This allows to effectively batch all the operations involved in the graph partitioning step, \ie, eigendecomposition of the Laplacian matrix and clustering, on all the graphs in the batch, regardless of their number of nodes.
Spectral clustering is applied on the sub-sampled graphs and the cluster assignments are propagated to the original graph: each node in the original graph is assigned the label of the temporally closest node in the subsampled~graph.

\subsection{Zero-shot procedural tasks implementation}
\ours can address several procedural tasks in zero-shot by framing them as a graph clustering problem.
We take graphs from different depths of the architecture depending on the task. For tasks that require video-language matching, such as step grounding or localization, we take the output of the last layer as the other layers are not language aligned.
For tasks where this constraint is not present, \eg, procedure learning on EgoProceL, we use features from deeper layers.
Clustering is computed using the Spectral Clustering implementation from \texttt{scikit-learn}.

\subsection{Features extraction with \ours}
On the Ego4D~\cite{ego4d} dataset, we utilize the official \texttt{omnivore\_video\_swinl} features and extract dense features from 16-frame windows with a stride of 16 frames using the EgoVLP~\cite{lin2022egocentric} and \textsc{LaViLa}~\cite{zhao2023learning} backbones. We follow the same procedure to extract features for the datasets in the EgoProceL~\cite{bansal2022my} benchmark.
When using \ours as a features extractor, \eg, to train VSLNet~\cite{zhang2020span} for the Step Grounding task, we take features from the output layer of the decoder. \ours's features have size 768 and maintain the same temporal granularity of the input features.

\section{Comparison between clustering algorithms}\label{sec:supp_exps}
\begin{table}
\centering
\scriptsize
\setlength{\tabcolsep}{5pt}
\begin{tabularx}{\columnwidth}{@{}Xccccc@{}}
\toprule
\multirow{2}{*}{\bf Features}& \multirow{2}{*}{\bf Algorithm} & \multicolumn{2}{c}{\bf mIoU@0.3} & \multicolumn{2}{c}{\bf mIoU@0.5}\\
\cmidrule(lr){3-4}\cmidrule(lr){5-6}
 & & R@1 & R@5 & R@1 & R@5 \\
\midrule
EgoVLP & KMeans (L2) & 10.37 & 24.65 & 6.85 & 16.46 \\
EgoVLP & KMeans (Cos.) & 8.97 & 23.21 & 5.91 & 15.15 \\
EgoVLP & Spectral & \underline{10.73} & 24.70 & \underline{7.38} & \underline{16.53} \\
\midrule
\textbf{Ours (EgoVLP)} & KMeans (L2) & 9.87 & 24.21 & 6.46 & 15.71 \\
\textbf{Ours (EgoVLP)} & KMeans (Cos.) & 10.35 & \underline{24.85} & 6.93 & 16.27 \\ 
\textbf{Ours (EgoVLP)} & Spectral & \textbf{11.57} & \textbf{27.41} & \textbf{7.87} & \textbf{18.70} \\
\bottomrule
\end{tabularx}
\vspace{-2.5mm}
\caption{\textbf{Impact of different clustering algorithms on the Step-Grounding task on Ego4D Goal-Step~\cite{song2024ego4d}}. We evaluate the baselines and \ours using KMeans and Spectral Clustering.}
\label{tab:supp_step-grounding-algo}
\end{table}

\begin{table*}[t]
\centering
\scriptsize
\setlength{\tabcolsep}{5pt}
\begin{tabularx}{\textwidth}{@{}Xccc|cccccccccccc@{}}
\toprule
\multirow{2}{*}{\bf Method} & \multirow{2}{*}{\bf Algorithm} & \multicolumn{2}{c}{\textbf{Average}} & \multicolumn{2}{c}{\textbf{CMU-MMAC}~\cite{cmu_mmac}} & \multicolumn{2}{c}{\textbf{EGTEA}~\cite{li2018eye}} & \multicolumn{2}{c}{\textbf{MECCANO}~\cite{ragusa_MECCANO_2023}} & \multicolumn{2}{c}{\textbf{EPIC-Tents}~\cite{epic_tent}} & \multicolumn{2}{c}{\textbf{PC Ass.}~\cite{bansal2022my}} & \multicolumn{2}{c}{\textbf{PC Disass.}~\cite{bansal2022my}} \\
\cmidrule(lr){3-4}\cmidrule(lr){5-6}\cmidrule(lr){7-8}\cmidrule(lr){9-10}\cmidrule(lr){11-12}\cmidrule(lr){13-14}\cmidrule(l){15-16}
\multicolumn{2}{c}{} &  F1 & \multicolumn{1}{c}{IoU} & F1 & IoU & F1 & IoU & F1 & IoU & F1 & IoU & F1 & \multicolumn{1}{c}{IoU} & F1 & IoU \\
\midrule
Omnivore & K-Means & 38.4 & 20.8 & 38.9 & 22.1 & 36.1 & 17.0 & 38.4 & 20.2 & 42.0 & 22.8 & 34.9 & 20.2 & 39.9 & 22.7 \\
Omnivore & Spectral & 39.1 & 22.0 & 44.7 & 26.8 & 37.1 & 19.2 & 36.0 & 19.0 & 40.8 & 21.9 & 35.7 & 21.5 & 40.3 & 23.5 \\
\midrule
EgoVLP & KMeans & 40.6 & 22.0 & 46.6 & 28.2 & 37.3 & 17.3 & 32.9 & 16.1 & 40.1 & 20.9 & 39.0 & 21.5 & 47.3 & 28.1  \\
EgoVLP & Spectral & 40.0 & 21.9 & 49.2 & \underline{31.0} & 36.6 & 18.3 & 33.1 & 16.1 & 37.4 & 19.2 & 38.2 & 20.8 & 45.4 & 25.6  \\
\midrule
\midrule
\textbf{Ours (Omnivore)} & K-Means & 43.7 & 24.2 & 46.9 & 27.3 & 38.6 & 18.4 & \textbf{43.9} & \textbf{24.4} & \underline{45.2} & \textbf{25.1} & 43.4 & 23.7 & 44.0 & 26.1 \\
\textbf{Ours (Omnivore)} & Spectral & 44.0 & 24.5 & 47.2 & 27.7 & \underline{39.7} & \textbf{19.9} & \underline{41.6} & \underline{22.1} & \textbf{45.3} & \underline{24.3} & 43.7 & \underline{25.1} & 46.3 & 27.9 \\
\midrule
\textbf{Ours (EgoVLP)} & K-Means & \underline{44.2} & \underline{24.7} & \underline{50.2} & 30.5 & \textbf{40.4} & \underline{19.8} & 39.5 & 20.4 & 41.8 & 22.2 & \underline{44.3} & 24.9 & \underline{48.9} & \underline{30.3} \\
\textbf{Ours (EgoVLP)} & Spectral & \textbf{44.5} & \textbf{25.3} & \textbf{53.5} & \textbf{34.0} & \underline{39.7} & 19.6 & 39.8 & 20.3 & 39.0 & 20.3 & \textbf{44.9} & \textbf{25.6} & \textbf{49.9} & \textbf{32.1} \\
\bottomrule
\end{tabularx}
\vspace{-2.5mm}
\caption{\textbf{Comparison of different clustering strategies on Omnivore and EgoVLP features~\cite{bansal2022my}}.}
\label{tab:egoprocel_algorithm}
\end{table*}

Our approach builds a similarity graph from the video segments and discovers functional threads as strongly connected regions of the graph. 
In this context, spectral clustering groups segments and actions that may not be close in terms of euclidean or cosine distance but are linked through similar actions, forming a strongly connected region of the graph.
We show the effectiveness of this design choice in Table~\ref{tab:supp_step-grounding-algo} on the Step Grounding task from Goal-Step~\cite{song2024ego4d}, comparing Spectral Clustering with KMeans using euclidean and cosine distances between the node embeddings. 
On the EgoVLP baseline, the two algorithms have similar performance.
Similarly, we evaluate different clustering algorithms on EgoProceL in Table~\ref{tab:egoprocel_algorithm}.

\section{Procedure step emergence in \ours}\label{sec:interpretability}
\begin{table}[t]
\centering
\scriptsize
\setlength{\tabcolsep}{5pt}
\begin{tabularx}{\columnwidth}{@{}Xcccc@{}}
\toprule
\multirow{2}{*}{\bf Method}& \multicolumn{2}{c}{\bf Zero-Shot} & \multicolumn{2}{c}{\bf Linear Probing}\\
\cmidrule(lr){2-3}\cmidrule(lr){4-5}
& Top-1 & Top-5 & Top-1 & Top-5 \\
\midrule
EgoVLP & \underline{10.11} & \underline{29.47}& \underline{25.22}& \underline{53.08}\\
\bf Ours (EgoVLP) & \bf 12.03& \bf 32.28 & \bf 30.22 & \bf 58.96\\
\bottomrule
\end{tabularx}
\vspace{-2.5mm}
\caption{\textbf{Key-step classification accuracy on Goal-Step~\cite{song2024ego4d}}, using an oracle for \emph{step} and \emph{substep} detection. Steps and substeps are more easily recognizable in the \ours feature space, despite no specific supervision.} 
\label{tab:goalstep_cls}
\end{table}

\begin{figure}[t]
\centering
\begin{subfigure}{0.5\columnwidth}
  \centering
  \includegraphics[width=1\columnwidth, trim=1.9cm 3.8cm 1.9cm 1.9cm,clip]{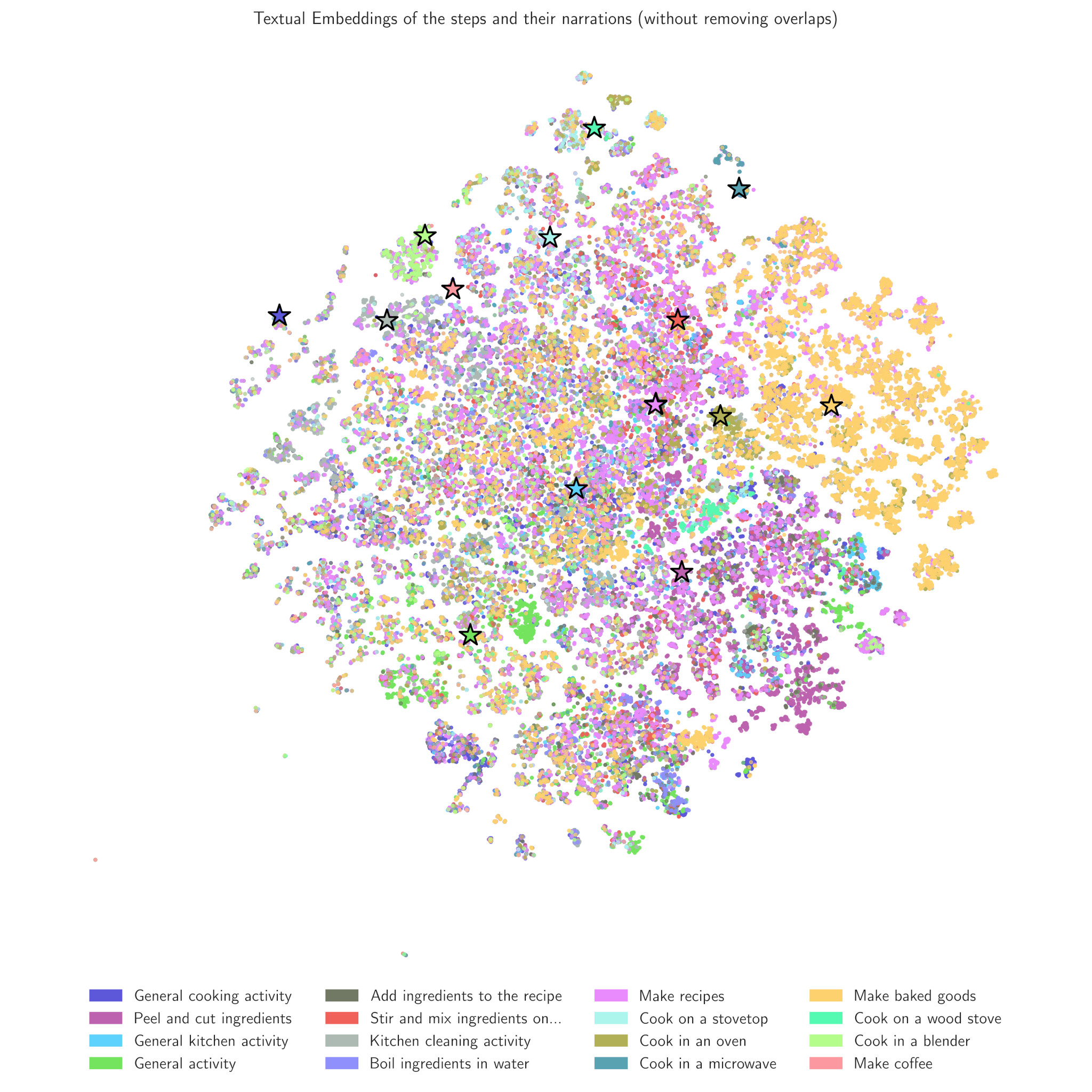}
  \caption{Ground truth steps}
  \label{fig:tsne_narrations_distribution_gt}
\end{subfigure}%
\begin{subfigure}{0.5\columnwidth}
  \centering
  \includegraphics[width=1\columnwidth, trim=1.9cm 3.8cm 1.9cm 1.9cm, clip]{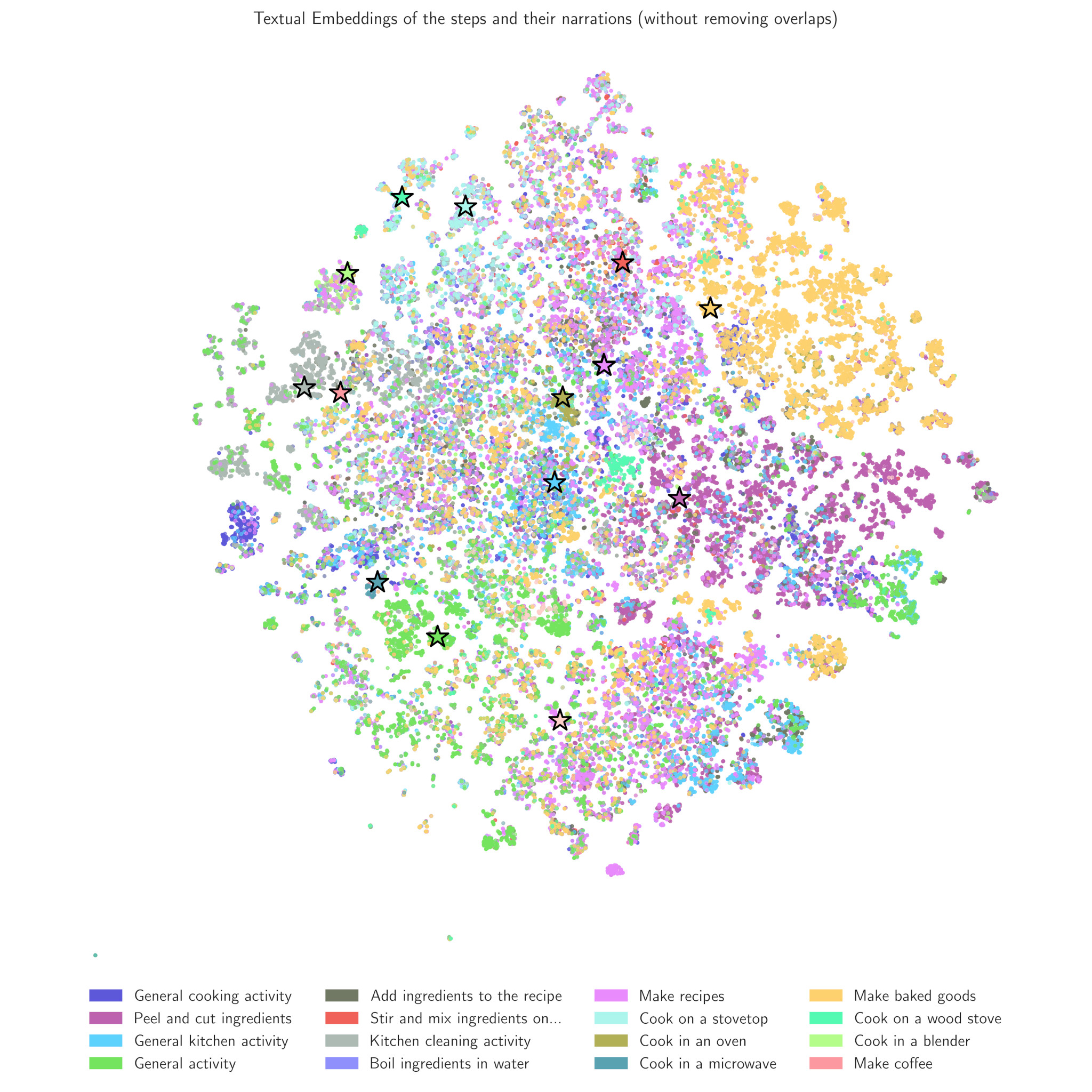}
  \caption{Predicted candidate steps}
  \label{fig:tsne_narrations_distribution_ours}
\end{subfigure}
\caption{\textbf{Features distribution of narrations and procedural steps in Goal-Step~\cite{song2024ego4d}.} Dots and stars represent the textual embeddings of the narrations and key-step labels, respectively, while the colors indicate the step to which the narrations are assigned. }
\label{fig:tsne_narrations_distribution}
\vspace{-\baselineskip}
\end{figure}

We evaluate the emergence of high-level \emph{functional threads} in \ours by analyzing the distribution of the textual embeddings for narrations and key-step labels from Goal-Step.
For each ground truth (Fig.~\ref{fig:tsne_narrations_distribution_gt}) or zero-shot step prediction (Fig.~\ref{fig:tsne_narrations_distribution_ours}), we collect all the narrations within the corresponding temporal window.
Our results show that \ours generates candidate steps where narrations are more tightly associated with the predicted key-step and form more distinct clusters, suggesting that narrations within the same step are semantically closer, irrespective of the granularity of the steps defined in the taxonomy.
To show that \ours features are more aligned with the key-step taxonomy despite no specific supervision, we train a linear probe on its features to predict the key-step label given the corresponding trimmed video segment (Table~\ref{tab:goalstep_cls}). 
Compared to EgoVLP, \ours improves noticeably the alignment between the visual features and the key-steps taxonomy ($\texttt{+}7.02\%$ top-1 accuracy), showing the steps and substeps are more easily recognizable in the \ours's feature space.

\begin{figure*}[t]
\centering
\begin{subfigure}{0.475\textwidth}
  \centering
  \includegraphics[width=1\columnwidth,trim={0 0 0 0.25cm},clip]{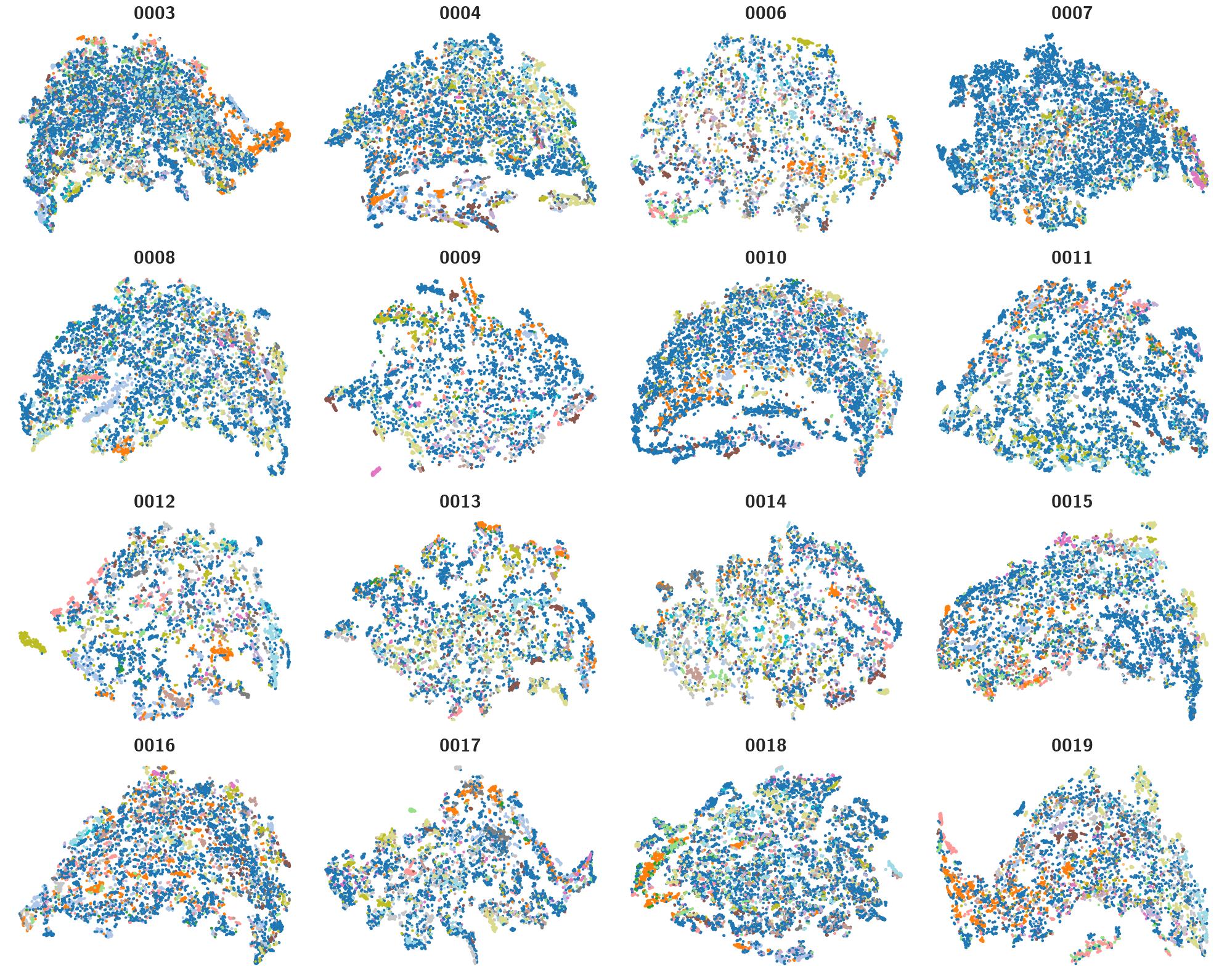}
  \caption{OPEL (Res-Net 50) Features}
  \label{fig:supp_meccano_tsne_opel}
\end{subfigure}%
\hfill
\begin{subfigure}{0.475\textwidth}
  \centering
  \includegraphics[width=1\columnwidth,trim={0 0 0 0.25cm},clip]{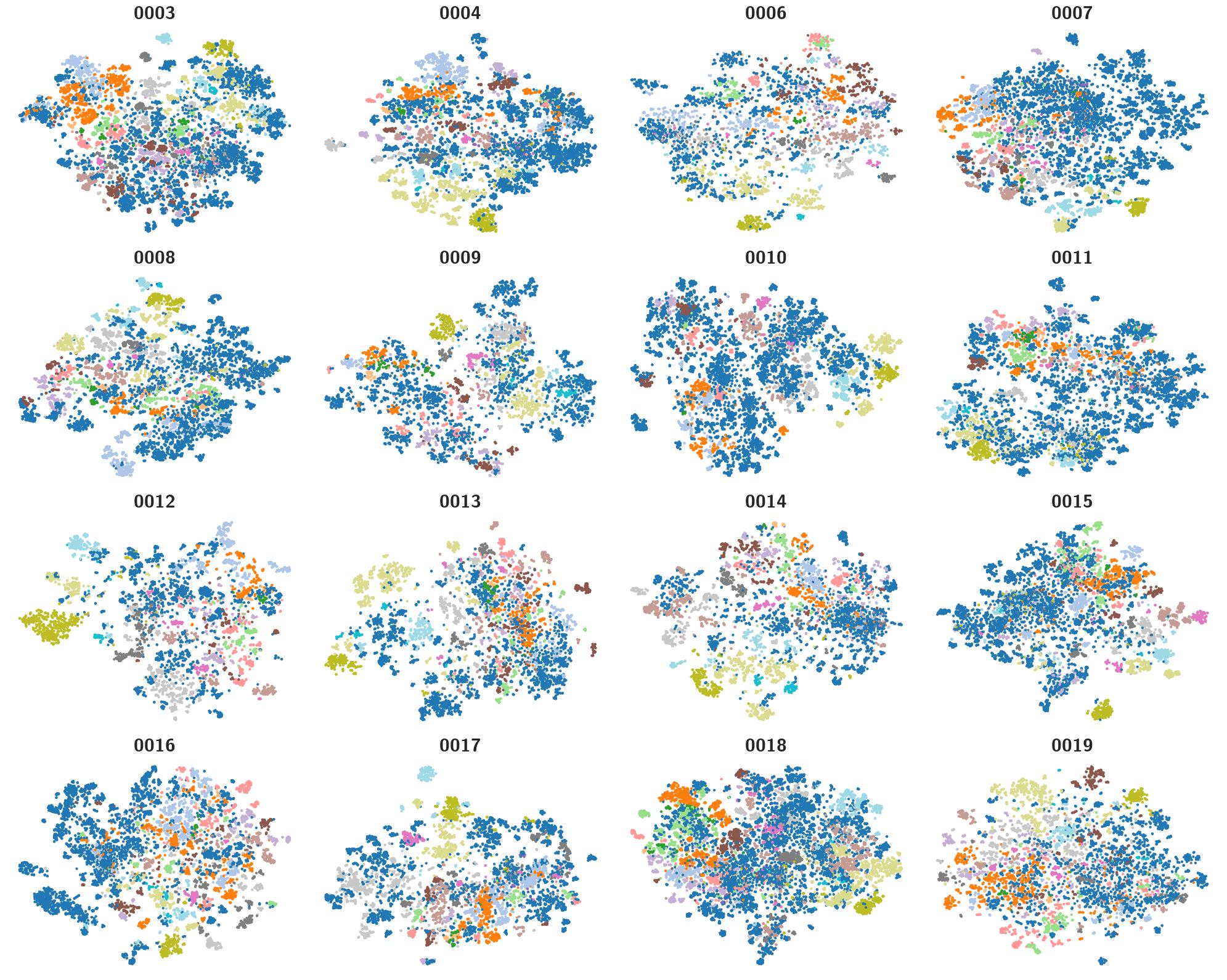}
  \caption{Omnivore Features}
  \label{fig:supp_meccano_tsne_omnivore}
\end{subfigure}
\caption{\textbf{Features distribution of Omnivore and OPEL on MECCANO~\cite{ragusa_MECCANO_2023}}, with dots representing different video segments, and colors encoding the ground truth step labels.
Despite not being trained on MECCANO, Omnivore features show a quite distinct separation between segments of the same action (same color). 
}
\vspace{-\baselineskip}
\label{fig:supp_meccano_tsne}
\end{figure*}
\section{OPEL with Omnivore backbone}\label{sec:opel_omnivore}
\begin{table}[t]
\centering
\scriptsize
\setlength{\tabcolsep}{3.5pt}
\begin{tabularx}{\columnwidth}{@{}Xccccccccccc@{}}
\toprule
& \multirow{2}{*}{\bf\shortstack{Context\\(Stride)}} & \multicolumn{2}{c}{\textbf{CMU}} & \multicolumn{2}{c}{\textbf{MEC.}} & \multicolumn{2}{c}{\textbf{PC Ass.}} & \multicolumn{2}{c}{\textbf{PC Dis.}} & \multicolumn{2}{c}{\textbf{Avg.}} \\
\cmidrule(lr){3-4}\cmidrule(lr){5-6}\cmidrule(lr){7-8}\cmidrule(lr){9-10}\cmidrule(lr){11-12}
& & F1 & IoU & F1 & IoU & F1 & IoU & F1 & \multicolumn{1}{c}{IoU} & F1 & IoU \\
\midrule
$-$ & 2 (15) & 36.5 & 18.8 & 39.2 & 20.2 & 33.7 & 17.9 & 32.2 & 16.9 & 35.4 & 18.5 \\
\midrule
OV & 2 (15) & 35.4 & 18.7 & 35.1 & 17.5 & 22.8 & 12.0 & 32.8 & 18.2 & 31.5 & 16.6 \\
OV & 4 (1)	& 31.6 & 20.1 & 36.9 & 18.3 & 33.0 & 18.8 & 31.0 & 16.4 & 33.1 & 18.4 \\
OV$^\dagger$ & 4 (1)	& 31.6 & 17.5 & 33.3 & 17.8 & 32.0 & 17.4 & 34.9 & 19.0 & 32.9 & 17.9 \\
\bottomrule
\end{tabularx}
\vspace{-2.5mm}
\caption{
\textbf{OPEL~\cite{chowdhuryopel} with Omnivore backbone}, comparing different temporal context windows. OV: Omnivore backbone. OV$^\dagger$: frozen Omnivore backbone.
}
\label{tab:opel_omnivore}
\end{table}

\noindent The Omnivore baseline significantly outperforms the previous SOTA on EgoProceL. 
We suggest that two main factors could explain the performance gap: (i) the different backbone and pre-training strategies used by OPEL (ResNet-50) and Omnivore, and (ii) different temporal contexts used for feature extraction.
We replace the ResNet-50 backbone in OPEL with Omnivore, varying the temporal context and stride used for features extraction (Table~\ref{tab:opel_omnivore}). 
The two backbones show comparable performance, with an improvement observed as the temporal context increases. We were unable to evaluate larger context windows due to memory overflows in the training process.
In addition, we show in Fig.~\ref{fig:supp_meccano_tsne} the features distribution of Omnivore against OPEL. 
Despite not being trained on MECCANO, Omnivore features exhibit quite clear clusters corresponding to the ground truth step labels. We argue that this behavior is the result of Omnivore being trained for action recognition on Kinetics-400.

\section{Additional visualizations}\label{sec:qualitatives}
\begin{figure*}[ht]
\centering
 \begin{subfigure}{.97\textwidth}
     \centering
     \includegraphics[height=0.75cm]{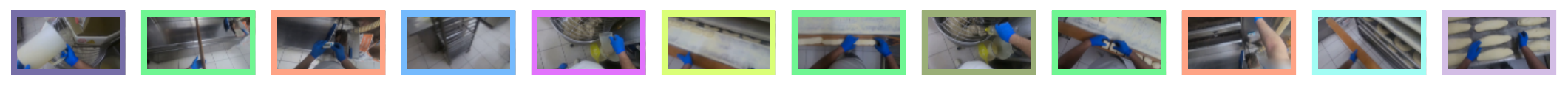}\\
     \includegraphics[width=.975\textwidth]{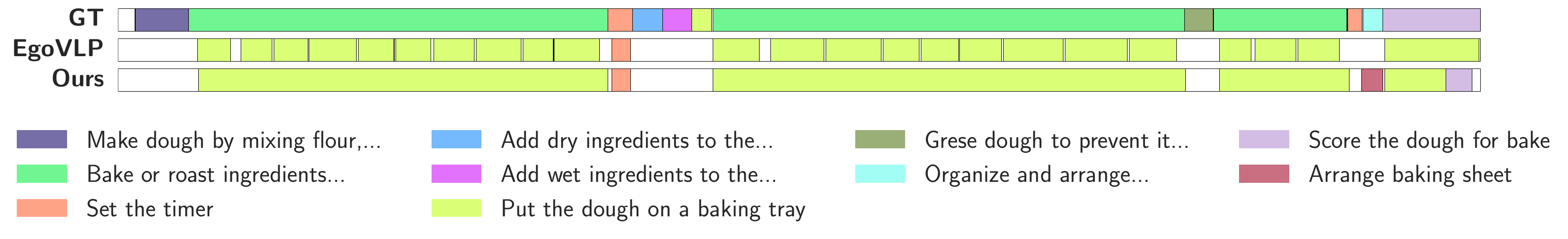}
     \caption{Video 1 (\texttt{2dbb7845-1de0-4e26-877a-035c051d12a5})}
     \label{fig:supp_zs_segm_a}
     \medskip
 \end{subfigure}
 \begin{subfigure}{.97\textwidth}
    \centering
    \includegraphics[height=0.75cm]{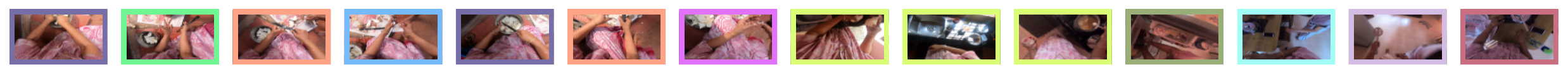}\\
    \includegraphics[width=.975\textwidth]{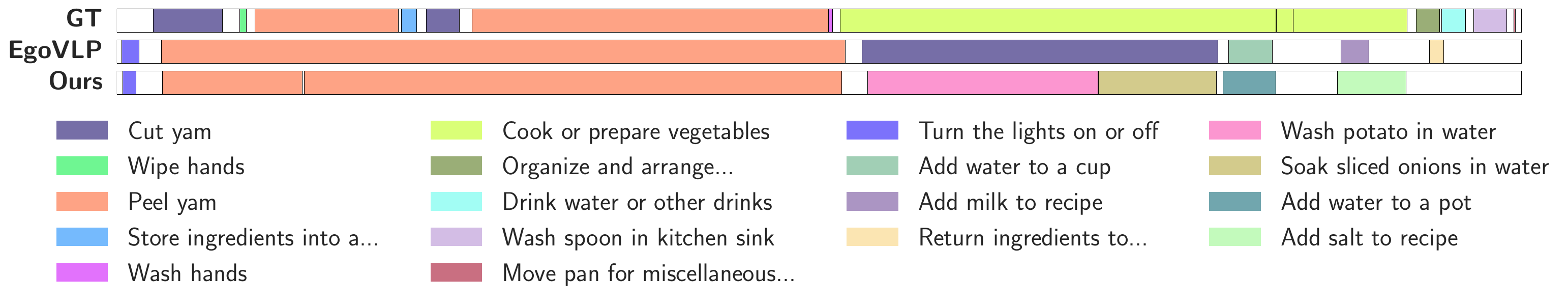}
    \caption{Video 2 (\texttt{f4cc5fdc-f64f-4dd7-9b95-61db9bbf33d5})}
    \label{fig:supp_zs_segm_b}
    \medskip
 \end{subfigure}
 \begin{subfigure}{.97\textwidth}
     \centering
     \includegraphics[height=0.75cm]{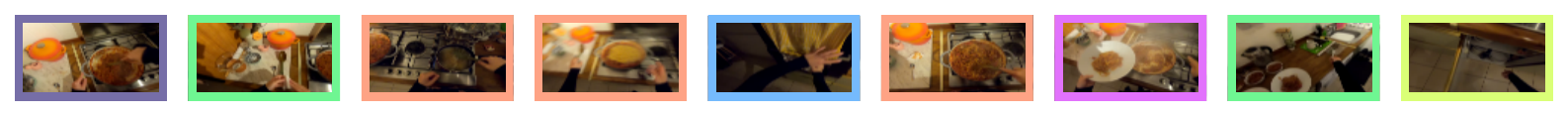}\\
     \includegraphics[width=.975\textwidth]{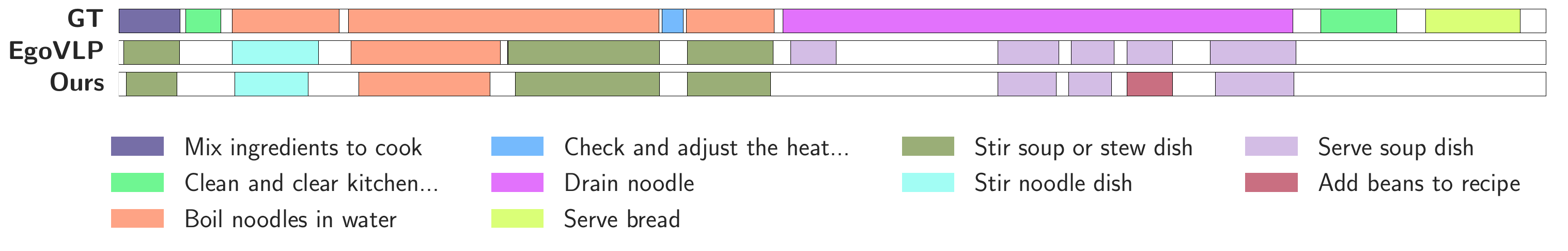}
     \caption{Video 3 (\texttt{c546c508-8352-4c5c-8770-e8f30fb4562a})}
     \label{fig:supp_zs_segm_c}
     \medskip
 \end{subfigure}
 \begin{subfigure}{0.97\textwidth}
     \centering
     \includegraphics[height=0.75cm]{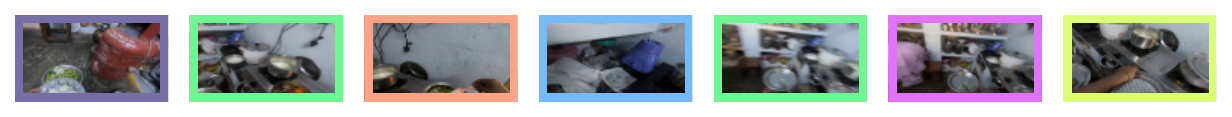}\\
     \includegraphics[width=.975\textwidth]{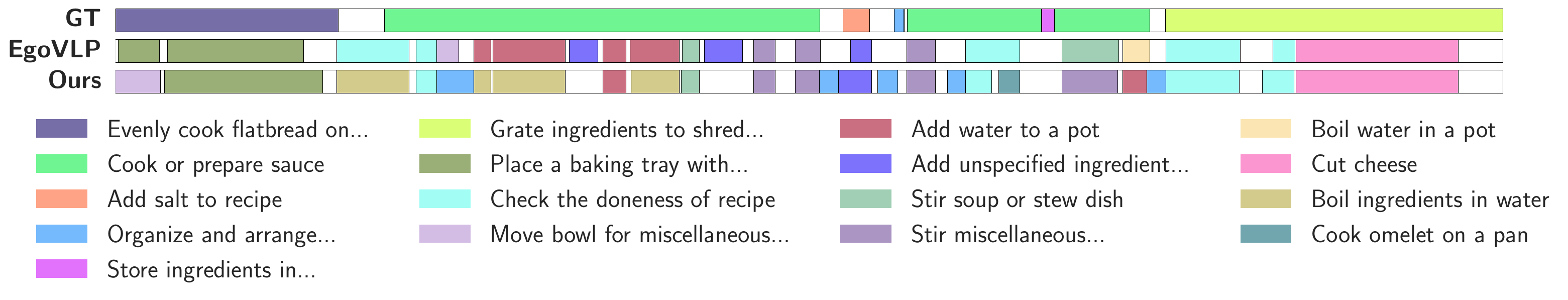}
     \caption{Video 4 (\texttt{acc6839e-9d6d-46db-921b-51812834d3b2})}
     \label{fig:supp_zs_segm_d}
 \end{subfigure}
 \caption{\textbf{Failure cases on the Zero-Shot Localization task on Goal-Step~\cite{song2024ego4d}}, showing the ground truth steps, the predictions obtained by clustering the EgoVLP and \ours features and the middle frame of each step from the ground truth.
 We find that most cases of failure are related to a mismatch between the granularity of ground truth steps and predictions. 
 In \textbf{Video 1} (Fig.~\ref{fig:supp_zs_segm_a}), both EgoVLP and \ours detect the most occurring step (\textcolor[RGB]{99,249,128}{$\blacksquare$} ``\emph{Bake or roast ingredients in oven}''), but EgoVLP is breaking the segment into more clusters and both methods confuse it with a similar step (\textcolor[RGB]{211,255,100}{$\blacksquare$} ``\emph{Put the dough on the baking tray}'').
 In \textbf{Video 2} (Fig.~\ref{fig:supp_zs_segm_b}), both EgoVLP and \ours group the initial part of the video in a single long step (\textcolor[RGB]{251,144,115}{$\blacksquare$} ``\emph{Peel yam}''). In the second half of the video, \ours predicts more fine-grained steps than the ground truth, \eg, (\textcolor[RGB]{249,126,197}{$\blacksquare$} ``\emph{Wash potato in water}'') rather than (\textcolor[RGB]{211,255,100}{$\blacksquare$} ``\emph{Cook or prepare vegetable}'').
 A similar issue appears in \textbf{Video 3} (Fig.~\ref{fig:supp_zs_segm_c}), in which there is a mismatch between the step ground truth, \eg, (\textcolor[RGB]{251,144,115}{$\blacksquare$} ``\emph{Boil noodles in water}'') and (\textcolor[RGB]{217,82,251}{$\blacksquare$} ``\emph{Drain noodle}'') and the predicted finer steps.
 \textbf{Video 4} (Fig.~\ref{fig:supp_zs_segm_d}) shows a more significant failure case where both methods predict many more steps than in the ground truth.
 }
 \label{fig:supp_zs_segm}

\end{figure*}

Fig.~\ref{fig:supp_zs_segm} shows additional qualitative results on the Step Localization task, comparing our approach with EgoVLP~\cite{lin2022egocentric}.
We observe that most failure cases are associated to mismatches between the temporal granularity of the predictions and the ground truth, or to confusion between semantically similar steps or sub-steps.

\end{document}